\documentclass{templates/arxiv/preprint}
\usepackage[utf8]{inputenc}
\usepackage{amsmath,amssymb,booktabs,tabularx,longtable,array,multirow}
\usepackage{graphicx,xcolor,url,hyperref}
\usepackage{placeins} %
\makeatletter\providecommand\insert@pcolumn{\insert@column}\makeatother
\usepackage{colortbl}
\usepackage{pifont} %
\definecolor{stepgray}{HTML}{3D4C61} %
\newcommand{\figstep}[1]{\textcolor{stepgray}{\ding{#1}}}
\ifdefined\pdfsuppressptexinfo\pdfsuppressptexinfo=-1\fi

\newcommand{\missing}[1]{\textbf{**}}
\newcommand{\system}{Design Anything}
\newcommand{\systemshort}{Design Anything}
\newcommand{\benchmark}{DesignAnything-Bench}
\newcommand{\systemtt}{\texttt{Design Anything}}
\newcommand{\benchmarktt}{\texttt{DesignAnything-Bench}}

\usepackage{etoolbox}
\definecolor{oursrow}{HTML}{EFE5F2} %

\newcommand{\tabfont}{\scriptsize\renewcommand{\arraystretch}{1.3}}
\newcommand{\tabtextfont}{\footnotesize\renewcommand{\arraystretch}{1.2}}
\newcommand{\tabhead}{\bfseries}
\newcommand{\oursname}[1]{\textbf{#1}}
\newcommand{\tabgroup}[1]{\textbf{#1}}
\newcommand{\papertableformat}[1]{}
\newcommand{\papertablecell}[2]{#1}
\newcommand{\papertablegainfont}{\scriptsize}
\newcommand{\papertableinterval}[3]{#1 [#2, #3]}

\newcolumntype{Y}{>{\raggedright\arraybackslash}X}

\newif\ifpapertwocolumn
\papertwocolumnfalse
\newcommand{\paperusesinglecolumnfloats}{%
  \let\paperfigure\figure\let\endpaperfigure\endfigure
  \let\papertable\table\let\endpapertable\endtable
}
\paperusesinglecolumnfloats
\newcommand{\paperusetwocolumnfloats}{%
  \renewenvironment{paperfigure}[1][tp]{\csname figure*\endcsname[tp]}{\csname endfigure*\endcsname}%
  \renewenvironment{papertable}[1][tp]{\csname table*\endcsname[tp]}{\csname endtable*\endcsname}%
}
\newcommand{\paperbibliographyfont}{}
\newcommand{\paperappendixsetup}{}
\newcommand{\paperbeforeconclusion}{}
\newif\ifpaperstatements
\paperstatementstrue

\paperstatementsfalse
\newcommand{\papertitle}{Design Anything, Design at Scale}

\definecolor{tsinghuapurple}{HTML}{660874}
\colorlet{tsinghualightpurple}{tsinghuapurple!4!white}
\renewcommand\authorformat[2][]{\mbox{\sffamily #2\textsuperscript{#1}}}
\renewcommand\affiliationformat[2][]{\mbox{\textsuperscript{#1}#2}}
\renewcommand\affiliation[2][]{\addtolist[#1]{#2}{\affiliationlist}{\affiliationformat}{\quad}}

\renewcommand{\tabhead}{\sffamily\bfseries}

\renewcommand{\mymaketitle}{%
  \begingroup
  \setlength{\parindent}{0pt}%
  \setlength{\parskip}{0pt}%
  \vspace*{-30pt}%
  {\noindent\includegraphics[height=32pt]{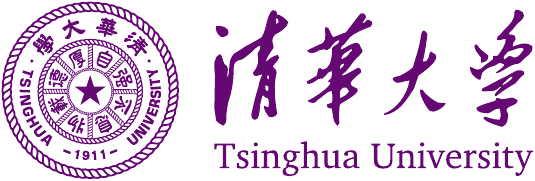}\par}%
  \vspace{5pt}%
  {\color{tsinghuapurple!45!white}\hrule height 0.4pt}%
  \vspace{14pt}%
  {\centering
    {\setstretch{1.2}\bfseries\titlelist\par}%
    \vspace{10pt}%
    \begin{minipage}{\linewidth}
      \centering\fontsize{9.5}{11}\selectfont\authorlist\par
    \end{minipage}\par
    \vspace{2pt}%
    {\fontsize{8.5}{10}\selectfont\affiliationlist\par}%
    \vspace{1pt}%
    {\fontsize{8}{9}\selectfont\contributionlist\par}%
  }%
  \vspace{13pt}%
  \begin{tcolorbox}[
    enhanced,
    colback=tsinghualightpurple,
    colframe=tsinghuapurple!25!white,
    boxrule=0.4pt,
    arc=6pt,
    left=0.45cm,right=0.45cm,
    top=0.24cm,bottom=0.24cm,
    before skip=0pt,after skip=0.2cm
  ]
    \abstractlist\par
    \printresources
    \ifdefempty{\metadatalist}{}{%
      \vspace{0.2cm}%
      \metadatalist\par
    }%
  \end{tcolorbox}
  \endgroup
  \FloatBarrier
}

\makeatletter
\gdef\@makecol{%
  \setbox\@outputbox\box\@cclv
  \let\@elt\relax
  \xdef\@freelist{\@freelist\@midlist}%
  \global\let\@midlist\@empty
  \@combinefloats
  \ifvoid\footins\else
    \setbox\@outputbox\vbox{%
      \boxmaxdepth\@maxdepth
      \unvbox\@outputbox
      \vskip\z@\@plus 1fill
      \vskip\skip\footins
      \color@begingroup
        \normalcolor
        \footnoterule
        \unvbox\footins
      \color@endgroup}%
  \fi
  \ifvbox\@kludgeins
    \@makespecialcolbox
  \else
    \setbox\@outputbox\vbox to\@colht{%
      \@texttop
      \dimen@\dp\@outputbox
      \unvbox\@outputbox
      \vskip-\dimen@
      \@textbottom}%
  \fi
  \global\maxdepth\@maxdepth}
\makeatother

\makeatletter
\edef\arxivrestorefloatlayout{%
  \noexpand\setlength{\noexpand\@fptop}{\the\@fptop}%
  \noexpand\setlength{\noexpand\@fpsep}{\the\@fpsep}%
  \noexpand\setlength{\noexpand\@fpbot}{\the\@fpbot}%
  \noexpand\renewcommand{\noexpand\topfraction}{\topfraction}%
  \noexpand\renewcommand{\noexpand\textfraction}{\textfraction}%
  \noexpand\renewcommand{\noexpand\floatpagefraction}{\floatpagefraction}%
}
\makeatother
\renewcommand{\topfraction}{0.9}
\renewcommand{\textfraction}{0.1}
\renewcommand{\floatpagefraction}{0.85}
\newcommand{\paperbibliographystyle}{templates/arxiv/plainnat}
\renewcommand{\paperappendixsetup}{%
  \arxivrestorefloatlayout
  \setkeys{Gin}{height=0.84\textheight,keepaspectratio}}
\title{\papertitle}
\author[1,*]{Xuchao Ren}
\author[1,*]{Siyuan Feng}
\author[1,2,*]{Jing Huang}
\author[1,3]{Yanhua Yao}
\author[1]{Jing Wang}
\author[1]{Lingyu Alina Wang}
\author[1]{Anan Chen}
\author[4]{Hangjie Yuan}
\author[1,\ensuremath{\dagger}]{Tao Feng}
\author[1,\ensuremath{\dagger}]{Qiong Wu}
\author[1,\ensuremath{\dagger}]{Sai Ma}

\affiliation[1]{Academy of Arts \& Design, Tsinghua University}
\affiliation[2]{Beijing Institute of Fashion Technology}
\affiliation[3]{Senseable City Lab, Massachusetts Institute of Technology}
\affiliation[4]{DAMO Academy, Alibaba Group}
\contribution[*]{Equal contribution}
\contribution[\ensuremath{\dagger}]{Corresponding authors}
\newcommand{\arxivlogo}[1]{\includegraphics[height=1.15em]{templates/arxiv/assets/#1}}
\definecolor{designanythingdark}{HTML}{18181B}
\newcommand{\arxivdesignanythinglogo}{\tikz[baseline=0pt]{%
  \fill[designanythingdark,rounded corners=0.25em] (0,0) rectangle (1.15em,1.15em);
  \node[text=white,font=\sffamily\bfseries\fontsize{4.5}{4.5}\selectfont,inner sep=0pt]
    at (0.575em,0.575em) {DA};}}
\resource{\arxivdesignanythinglogo}{Project page}{https://designanything.gemus.ai}
\resource{\arxivlogo{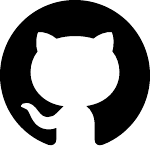}}{Benchmark and evaluator}{https://github.com/Gemus-AI/DesignAnything}
\resource{\arxivlogo{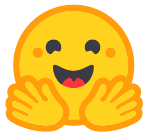}}{Dataset}{https://huggingface.co/datasets/Gemus-AI/DesignAnything-Bench}
\resource{\arxivlogo{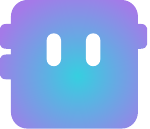}}{Gemus platform}{https://gemus.ai/community/collections/design-anything}

\newcommand{\arxivmailicon}{\tikz[baseline=0pt,line width=0.4pt,line join=round]{%
  \draw (0,0) rectangle (0.66em,0.48em);
  \draw (0,0.48em) -- (0.33em,0.18em) -- (0.66em,0.48em);}}
\newcommand{\arxivmail}[1]{\mbox{\arxivmailicon\,\email{#1}}}
\makeatletter
\newcommand{\arxivcorrespondencenote}{{\let\thefootnote\relax
  \renewcommand\@makefntext[1]{\noindent##1}%
  \footnotetext{\raggedright\textsf{\textbf{Contact:}}\hspace{0.6em}%
    \arxivmail{rechaos@mail.tsinghua.edu.cn}\hspace{1em}%
    \arxivmail{fengsy24@mails.tsinghua.edu.cn}\hspace{1em}%
    \arxivmail{fengtao.hi@gmail.com}%
    }}}
\makeatother

\abstract{As multimodal resources are transformed into media outputs, agentic AI is expected to stay aligned with design intent over long horizons. Designers follow briefs specifying goals, constraints, and deliverables, with later work building on earlier decisions. Recent work reuses design experience across tasks, yet alignment among artifacts within one project remains challenging. Agents therefore need to preserve and apply this intent across domains and as projects grow. We introduce \systemtt, a graph-based harness that preserves design intent outside the model and binds deliverables to the workflow steps that produce them through an intent--execution graph. A task contract retains constraints and decisions, selected designs become explicit inputs to later generations, and a deterministic controller uses bindings and execution records to check output types and counts, identify outstanding deliverables, and guide subsequent turns. We also introduce \benchmarktt, with 228 tasks across 19 design domains, including 38 project families at scales of 6, 14, and 24 artifacts. With the LLM held fixed, \systemtt\ satisfies 85.0\% and 92.5\% of the brief's factual constraints with DeepSeek V4.1 Flash High and GPT-5.6 Sol High, respectively, exceeding the strongest same-model baselines by 15.4\% and 14.4\%. These gains reach 25.1\% and 23.0\% on 24-artifact tasks. Under each model, \systemtt\ leads or ties in constraint satisfaction in 18 of 19 domains. Professional designers also prefer its complete deliverable sets over those of same-model baselines in a blind evaluation.
}

\begin{document}
\maketitle
\newcommand{\introgainfigure}[1][\linewidth]{%
\begin{paperfigure}[!b]
\centering
\includegraphics[width=#1]{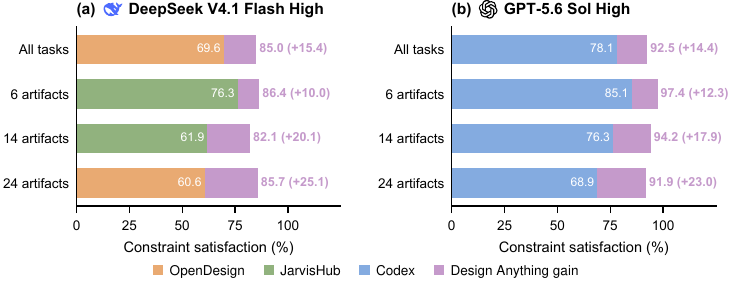}
\caption{\textbf{\system\ outperforms the best same-model baseline at every scale.} Colored bars show the strongest baseline and purple segments the gain of \system; labels give its rate and lead in percentage points.}
\label{fig:gain}
\end{paperfigure}}
\ifdefined\arxivcorrespondencenote
\introgainfigure[0.94\linewidth]
\arxivcorrespondencenote
\section{Introduction}
\label{sec:introduction}
\else
\section{Introduction}
\label{sec:introduction}
\introgainfigure
\fi

Human design activity often begins with a specific need, such as developing a product, staging an exhibition, or building a brand. Designers work from a design brief, which states the initial design intent: the project's goals, constraints, and required deliverables. Such projects usually call for a set of coordinated artifacts, including concept images, renderings, specification documents, videos, and slide decks, and later work builds on earlier design decisions, such as the selected concept and references~\citep{gero1990designprototypes,gero2004situated,dorst2001creativity}.

\begin{paperfigure}[t]
  \centering
  \includegraphics[width=\linewidth]{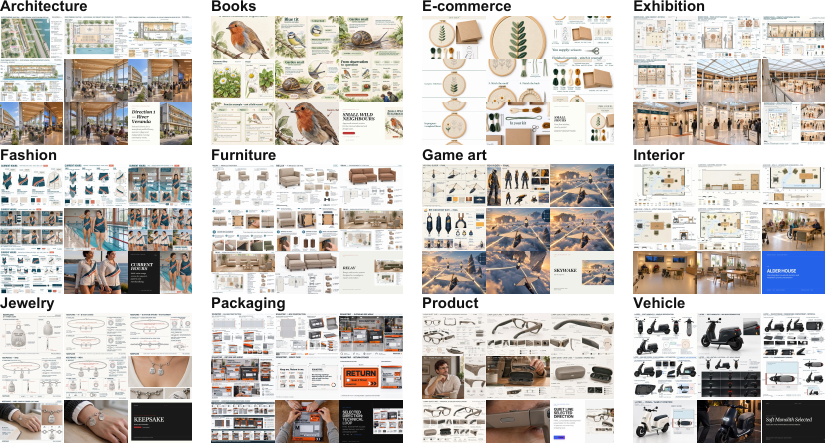}
  \caption{\textbf{Design anything across twelve design domains.} Each group shows artifacts from a single run of \system\ with GPT-5.6 Sol High, using the same configuration for every domain, without skills written for a particular design domain or task-specific hints.
  }
  \label{fig:teaser}
\end{paperfigure}

Generative models and design tools can now produce individual artifacts~\citep{openai2026images2,bytedanceseed2026seedance2,tripo3d,zheng2025pptagent,ge2025autopresent}. Recent work on agentic artifact creation highlights the challenge of preserving requirements and earlier decisions throughout multi-step production~\citep{wang2026agenticartifact}. Long-horizon tasks remain difficult~\citep{kwa2025measuring,liu2024agentbench,xie2024osworld}. Models do not reliably use the information in long contexts~\citep{liu2024lost,hong2025context}, perform worse when requirements arrive over multiple turns~\citep{laban2026lost}, and, in long tool-use runs, stop too early, verify too little, and lose constraints~\citep{yuan2026osworld2,merrill2026terminalbench}. Staying aligned with design intent means delivering every requested artifact, meeting each constraint in every artifact it concerns, and carrying earlier design decisions forward so that the artifacts stay consistent; all three become harder as deliverables multiply. An agent's performance depends not only on its large language model (LLM) but also on its harness, the software layer that manages tool calls, context, and multi-step execution~\citep{yao2023react,shinn2023reflexion,lee2026metaharness,wu2026harnessdev,yang2024sweagent,agashe2025agents,autodesign2026}.

Existing studies have explored harness mechanisms for sustained execution, including persistent requirement records and independent review of completion status~\citep{ma2026longhorizon}, as well as canvases that organize artifacts and their reference relationships~\citep{jarvishub2026}. Other studies reuse design experience accumulated over earlier tasks~\citep{autodesign2026}. These approaches either rely on a model to judge whether the work is complete or check one artifact at a time. Graph-based agent orchestration, recently discussed as graph engineering, makes task dependencies and execution state explicit~\citep{kim2024llmcompiler,wei2026graphharness,feng2026graphengineering}. For long-horizon design, the challenge for a harness is to ensure that recorded constraints and design decisions continue to guide artifact production and to reliably identify unfinished deliverables. Therefore, this raises a natural question: \emph{how can agentic AI consistently align with design intent over long-horizon tasks?}

To this end, we propose \systemtt, a graph-based harness for long-horizon agentic design whose key idea is to keep design intent outside the model, in an explicit state the harness can check, and to bind each deliverable to the workflow steps that produce it; we call this state the \emph{intent--execution graph} (Figure~\ref{fig:architecture}). A persistent task contract records the deliverables, constraints, and design decisions made during the run, so later turns still see them. When a later artifact builds on a selected concept or a supplied reference, the agent links it into the graph as an explicit input, so the artifact is generated from it rather than from a fresh description (Figure~\ref{fig:luma-case}). After each turn, a deterministic controller uses the bindings, the execution records, and checks on output kind and count to find the outstanding deliverables, and starts another turn that names them; a run succeeds only when none remain. The same configuration serves every domain (Figure~\ref{fig:teaser}; Appendix~\ref{app:cases}).

Evaluating such a harness requires a suitable benchmark. Some existing benchmarks include professional tasks with several deliverables~\citep{sun2026agentslastexam,li2026weavebench}, among them design and creative tasks~\citep{li2026weavebench}; others focus on workflow construction, tool planning, UI reproduction, poster generation, or single-image quality~\citep{xue2024comfybench,ki2025graphicbench,jeong2025canvas,si2025design2code,pang2025paper2poster,li2026qwenimagebench}. However, results at one project scale, the number of requested artifacts, do not show whether an agent stays aligned with design intent as the same project grows. We introduce \benchmarktt\ for this purpose, with 228 tasks across 19 design domains, including 38 project families each posed at 6, 14, and 24 requested artifacts with shared background and inputs. \systemtt\ satisfies 85.0\% of the brief's factual constraints (such as names and dimensions) with DeepSeek V4.1 Flash High and 92.5\% with GPT-5.6 Sol High. Against the strongest baseline on the same LLM, it leads by 15.4\% and 14.4\% over all tasks and by 25.1\% and 23.0\% on 24-artifact tasks (Figure~\ref{fig:gain}), and it leads or ties in 18 of 19 domains under each model (Figure~\ref{fig:domains}). In a blind evaluation of complete output sets by professional designers, the two \systemtt\ configurations receive the two highest estimated preference scores among the five configurations. Our contributions are threefold:
\begin{itemize}
  \item \textbf{Design Anything Harness.} We propose \systemtt, a harness that keeps design intent outside the model in an intent--execution graph binding each deliverable to the workflow steps that produce it.
  \item \textbf{DesignAnything-Bench.} We introduce \benchmarktt, a 228-task benchmark spanning 19 design domains, with 38 project families scaled from 6 to 24 artifacts, to test whether agents stay aligned with design intent across domains and as projects grow.
  \item \textbf{Evaluation.} With the LLM fixed, \systemtt\ satisfies more constraints than the strongest baseline overall and at every project scale, leads or ties in 18 of 19 domains, and receives the highest designer preference.
\end{itemize}

\section{Method}
\label{sec:method}

\system\ keeps design intent in an explicit state outside the model and operates on it with two loops (Figure~\ref{fig:architecture}). After formalizing what alignment with design intent requires of a run (\figstep{182}), we describe this state, the intent--execution graph (\figstep{183}), which binds each deliverable to the workflow steps that produce it. The agent loop (\figstep{184}) builds and executes this graph, and the delivery loop (\figstep{185}) checks it to find unfinished deliverables and decide the next action.

\begin{paperfigure}[t]
  \centering
  \ifdim\linewidth>430pt
    \includegraphics[width=\linewidth]{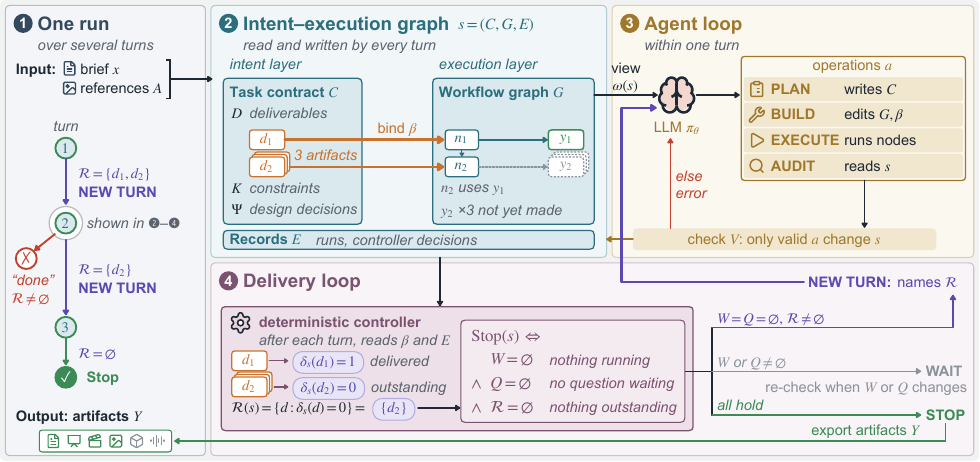}
  \else
    \includegraphics[width=\linewidth]{figures/design-anything-v64-layers.pdf}
  \fi
  \caption{\textbf{Overview of Design Anything.} Within a run (\figstep{182}), the intent--execution graph (\figstep{183}) records design intent and binds deliverables to workflow nodes. The agent loop (\figstep{184}) builds and executes the workflow; the delivery loop (\figstep{185}) uses execution records to identify outstanding deliverables and decide the next action.}
  \label{fig:architecture}
\end{paperfigure}

\subsection{Problem formulation}
\label{sec:method-task}

Given a design brief $x$ and reference materials $A$, we hold the LLM $\pi_\theta$ fixed and write the artifacts exported by \system\ as $Y\sim H(\pi_\theta,x,A)$. The brief states the project's goals, constraints $K$, and deliverables $D=\{d_1,d_2,\dots\}$. Several deliverables may share a selected concept, as with product views, a specification document, and a slide deck. In this setting, staying aligned with design intent requires a run (\figstep{182} in Figure~\ref{fig:architecture}) to produce an artifact for every $d\in D$, meet each constraint in every deliverable it concerns, and carry earlier design decisions into later artifacts. These properties are measured from exported artifacts (Section~\ref{sec:benchmark-validity}). For constraint satisfaction, the expected task-level score is
\begin{equation}
J_{\mathrm{CS}}(H)=\mathbb{E}_{(x,A)\sim p_{\text{task}},\;Y\sim H(\pi_\theta,x,A)}\big[\mathrm{CS}(Y;x)\big],
\label{eq:objective}
\end{equation}
where $p_{\text{task}}$ is the task distribution and $\mathrm{CS}(Y;x)$ is defined in Appendix~\ref{app:measures}. As deliverables multiply, each turn needs the relevant constraints and earlier decisions, together with the current delivery status.

\subsection{Intent--execution graph}
\label{sec:method-contract}
\label{sec:method-workflow}

The intent--execution graph (\figstep{183} in Figure~\ref{fig:architecture}) stores the task contract $C$, workflow graph $G$ with nodes $N_G$, and execution records $E$:
\begin{equation}
s=(C,G,E),\qquad C=(D,K,\Psi),\qquad \beta: D\to 2^{N_G}.
\label{eq:state}
\end{equation}
Here $\Psi$ contains recorded design decisions, and $\beta$ binds deliverables to their producing nodes.

The task contract $C$ keeps deliverables, constraints, and recorded design decisions available across turns, together with goals, success criteria, and open questions. The same contract format and delivery rules apply across domains; domain-specific constraints come from the brief.

To carry selected designs into later artifacts, the execution layer represents their reuse as explicit data dependencies in $G$, as in graph-based agent orchestration~\citep{wei2026graphharness,feng2026graphengineering}. $G$ is a typed directed acyclic graph that only the agent edits during a run, through tool calls (Section~\ref{sec:method-agent}). Edges connect compatible output and input ports. Each node $n$, such as $n_1$ producing $y_1$ in Figure~\ref{fig:architecture}, computes its output from its configuration $p_n$ and the outputs of its parent nodes $\mathrm{pa}(n)$:
\begin{equation}
y_n=g_n\big(p_n,\{y_{n'}:n'\in\mathrm{pa}(n)\}\big).
\label{eq:graph}
\end{equation}
A selected concept or supplied reference connected this way enters later generation as an explicit input, rather than being reconstructed from a fresh description. In Figure~\ref{fig:architecture}, node $n_2$ is configured to use $n_1$'s output $y_1$ as an input; Figure~\ref{fig:luma-case} illustrates such connections in a recorded design project. Appendix~\ref{app:reference-coverage} measures how often generations receive reference inputs.

To determine unfinished work, the binding $\beta$ connects deliverables in $C$ to their producing nodes in $G$. Together with $E$, which records node states and controller decisions, it lets the delivery loop derive outstanding deliverables (Section~\ref{sec:method-control}). Appendix~\ref{app:method-details} gives the implementation details.

\subsection{Agent loop}
\label{sec:method-agent}

The agent loop builds and revises this state through checked tool operations (\figstep{184} in Figure~\ref{fig:architecture}). Each turn lasts until the model stops calling tools. The model observes $\omega(s)$, comprising the canvas that renders $G$ and its artifacts and a state summary regenerated between turns.

The model selects operations $a\sim\pi_\theta(\cdot\mid x,A,\omega(s))$ from four tool families, in any order: PLAN records or revises intent in $C$; BUILD creates or edits nodes, edges, and deliverable bindings in $G$; EXECUTE runs nodes; and AUDIT reads $C$, $G$, and $E$. An operation takes effect only if a deterministic validity check $V$ accepts it:
\begin{equation}
s'=\begin{cases}
\Phi(s,a) & \text{if } V(s,a),\\
s & \text{otherwise,}
\end{cases}
\label{eq:transition}
\end{equation}
where $\Phi$ is the operation's state transition. $V$ checks tool arguments, node configurations, references, port types, and bindings; a rejected operation leaves the state unchanged and returns an actionable error.

\subsection{Delivery loop}
\label{sec:method-control}

The delivery loop uses the bindings to identify unfinished deliverables when the model ends a turn (\figstep{185} in Figure~\ref{fig:architecture}). A deterministic controller computes delivery status from $G$ and $E$:
\begin{equation}
\ifpapertwocolumn\begin{gathered}\fi
\delta_s(d)=\mathbf{1}\big[\beta(d)\neq\emptyset\,\wedge\,\mathrm{settled}_s(\beta(d))\,\wedge\,\mathrm{ok}_s(d)\big],\ifpapertwocolumn\\\else\qquad\fi
\mathcal{R}(s)=\{d\in D:\delta_s(d)=0\},
\ifpapertwocolumn\end{gathered}\fi
\label{eq:status}
\end{equation}
Here $\mathrm{settled}_s(\beta(d))$ requires every bound executable node to have finished executing, and $\mathrm{ok}_s(d)$ requires the outputs counted for $d$ to belong to that deliverable and pass the checks on output kind and item count. Thus $\mathcal{R}(s)$ includes both unbound deliverables and those whose bound work remains unfinished, such as $d_2$ before its node $n_2$ produces $y_2$ in Figure~\ref{fig:architecture}.

To direct subsequent work, let $W(s)$ contain running jobs and delegated tasks, and $Q(s)$ contain questions and decisions awaiting input. The controller also runs when a job finishes or an awaited input arrives. A run ends successfully only when
\begin{equation}
\mathrm{Stop}(s)\iff W(s)=\emptyset\,\wedge\,Q(s)=\emptyset\,\wedge\,\mathcal{R}(s)=\emptyset.
\label{eq:stop}
\end{equation}
A model's statement that it has finished is therefore insufficient (\figstep{182} in Figure~\ref{fig:architecture}). After applying failure checks, the controller waits while $W$ or $Q$ is non-empty, ends the run when $\mathrm{Stop}$ holds, and otherwise starts a new turn naming the outstanding deliverables and their blocking reasons. Failures or lack of progress can instead end the run unsuccessfully; Appendix~\ref{app:method-details} specifies the control order and stopping rules.

The controller checks delivery using declared node capabilities and output item counts. Constraint satisfaction and consistency are evaluated from the exported artifacts (Section~\ref{sec:experiments}). We also compare the full harness with reduced execution configurations to assess delivery and efficiency (Section~\ref{sec:components}).

\section{Experiments}
\label{sec:experiments}
Holding the LLM fixed, we ask: (1) Can one harness fulfill briefs across domains (\emph{Anything})? (2) Can it do so as projects require more artifacts (\emph{At Scale})? (3) How does the full harness compare with reduced configurations in delivery and efficiency? We also ask which delivered sets professional designers prefer.

\subsection{Benchmark: \benchmark}
\label{sec:benchmark}
\label{sec:benchmark-construction}
\label{sec:benchmark-scale}
\label{sec:benchmark-metrics}

\subsubsection{Task composition}
The benchmark spans 19 design domains with 12 tasks each (Figure~\ref{fig:benchmark-overview}). Tasks supply English briefs, input materials, and delivery requirements. For matched-scale comparisons, 38 project families share background and inputs across 6-, 14-, and 24-artifact versions (114 tasks). The 57 reference-material tasks test matching supplied objects; 57 breadth tasks add subjects and audiences. Each of these independent projects requests 6 or 14 artifacts. In total, 228 tasks represent 152 distinct projects (Appendix~\ref{app:benchmark-protocol-details}).

\begin{paperfigure}[!htbp]
\centering
\includegraphics[width=\linewidth]{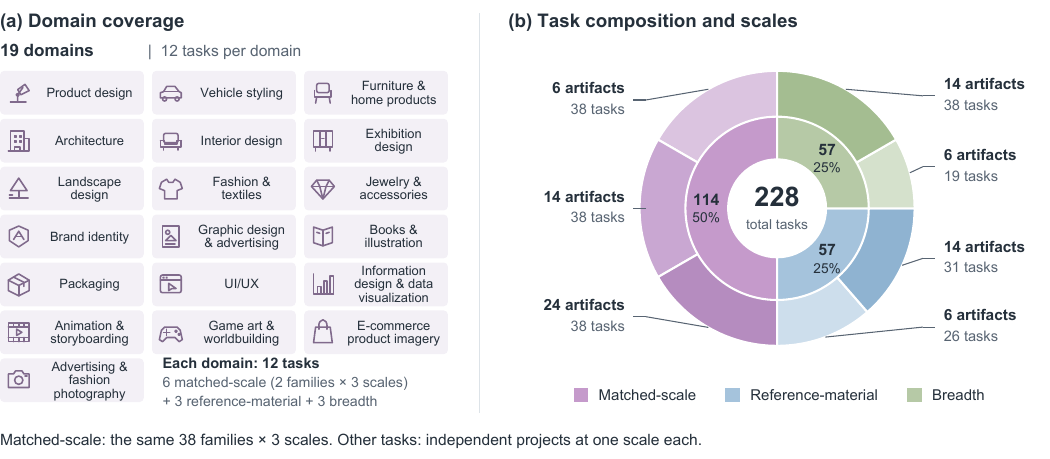}
\caption{\textbf{\benchmark\ contains 228 tasks across 19 domains.} In (b), the inner ring shows task conditions; the outer ring shows requested artifact counts. Only matched-scale tasks reuse the same 38 families across scales; each other task is an independent project at one scale. Artifact count measures scale.}
\label{fig:benchmark-overview}
\end{paperfigure}

\subsubsection{Evaluation measures}
\label{sec:benchmark-validity}
In our work, delivery completion, constraint satisfaction, and cross-artifact consistency measure observable aspects of alignment in exported artifacts. Automatic evaluation follows Agent-as-a-Judge~\citep{zhuge2025agentasajudge} (Appendix Table~\ref{tab:measures}).

\emph{Requirement fulfillment} measures how well outputs meet the brief. Delivery completion is the share of requested files delivered, capped per category. Constraint satisfaction is the share of applicable factual constraints satisfied in every eligible artifact. For illustration, delivering three requested posters gives 100\% delivery, but a name required on all three and correct on only two fails that constraint. Table~\ref{tab:main-results} pools satisfied constraints, not fully successful tasks. Reference fidelity measures the share of output--reference pairs depicting the same object (Appendix~\ref{app:measures}).

\emph{Cross-artifact consistency} measures agreement among outputs. Identity consistency compares outputs with a designated output from the same run (the anchor), measuring the share of pairs depicting the same object. Reference fidelity instead uses a supplied asset. Set consistency scores shared visual rules, such as palette and typography, on a 100-point rubric.

\emph{Design quality} (100-point scale) covers technical quality, aesthetics, text rendering, and domain criteria~\citep{li2026qwenimagebench}. Aggregate image-checklist means are multiplied by aggregate delivery completion, reflecting both image quality and missing work (Appendix~\ref{app:measures}). A vision-language judge scores the checklists; Appendix~\ref{app:judge-calibration} documents calibration of the other automatic judges against human annotations.

\emph{Designer preference} asks professional designers which delivered set they would hand to the client, comparing outputs for the same brief without seeing system names and allowing ties. Table~\ref{tab:main-results} summarizes these choices using Bradley--Terry estimates (Section~\ref{sec:human}).

\subsection{Experimental setup}
\label{sec:experimental-setup}

\paragraph{Systems.}
In the main comparison, a configuration pairs a system with an LLM. We group configurations by LLM. On DeepSeek V4.1 Flash High, we compare \system\ with the design agents OpenDesign~\citep{opendesign2026} and JarvisHub~\citep{jarvishub2026}; on GPT-5.6 Sol High, we compare it with Codex CLI. All systems generate images with gpt-image-2.5-sunburst. These comparisons evaluate each complete system with its own tools and execution policies (Appendix Table~\ref{tab:tracks}).

\paragraph{Protocol.}
All systems receive the same briefs and materials except on three tasks of one book project family (Appendix~\ref{app:benchmark-protocol-details}), without skills written for a particular design domain or task-specific hints. Each task contributes one final result per configuration, scored by a common evaluator. We compare scores on the same tasks and estimate uncertainty using 95\% bootstrap intervals over project families (Appendix~\ref{app:robustness}).

\subsection{Benchmark results}
\label{sec:overall}

\subsubsection{Overall and cross-domain results}

\begin{papertable}[t]
\centering\tabfont
\setlength{\tabcolsep}{1pt}
\papertableformat{main-results}
\newcommand{\modelgroup}[1]{\multicolumn{11}{l}{#1}}
\newcommand{\ours}{\rowcolor{oursrow}}
\newcolumntype{M}[2]{>{\hsize=\papertablecell{#1}{#2}\hsize\linewidth=\hsize\centering\arraybackslash}X}
\caption{\textbf{Results on \benchmark\ (228 tasks).} Fulfillment, identity consistency, and designer preference are percentages; other scores use a 100-point scale. Higher is better; bold marks the within-model best. Design quality is weighted by delivery completion; designer preference uses Bradley--Terry estimates (Section~\ref{sec:human}).}
\label{tab:main-results}
\begin{tabularx}{\linewidth}{@{}lM{1.05}{1.20}M{1.10}{0.94}M{1.00}{1.42}M{1.10}{1.13}M{1.10}{0.56}M{0.90}{0.88}M{1.00}{0.95}M{0.90}{0.72}M{0.85}{1.13}M{1.00}{1.07}@{}}
\toprule
& \multicolumn{3}{c}{\tabhead Requirement fulfillment} & \multicolumn{2}{c}{\tabhead Consistency} & \multicolumn{4}{c}{\tabhead Design quality} & {\tabhead Human} \\
\cmidrule(lr){2-4}\cmidrule(lr){5-6}\cmidrule(lr){7-10}\cmidrule(l){11-11}
{\tabhead System} & {\tabhead \papertablecell{\shortstack{Delivery\\completion}}{Delivery}} & {\tabhead \papertablecell{\shortstack{Constraint\\satisfaction}}{\shortstack{Con-\\straint}}} & {\tabhead \papertablecell{\shortstack{Reference\\fidelity}}{Reference}} & {\tabhead \papertablecell{\shortstack{Identity\\consistency}}{Identity}} & {\tabhead \papertablecell{\shortstack{Set\\consistency}}{Set}} & {\tabhead \papertablecell{\shortstack{Technical\\quality}}{\shortstack{Tech-\\nical}}} & {\tabhead \papertablecell{Aesthetics}{\shortstack{Aes-\\thetics}}} & {\tabhead \papertablecell{\shortstack{Text\\rendering}}{Text}} & {\tabhead \papertablecell{\shortstack{Domain\\criteria}}{Domain}} & {\tabhead \papertablecell{\shortstack{Designer\\preference}}{\shortstack{Pref-\\erence}}} \\
\midrule
\modelgroup{\tabgroup{(a) DeepSeek V4.1 Flash High}} \\
\ours \oursname{\system} & \textbf{96.6} & \textbf{85.0} & \textbf{75.6} & \textbf{84.5} & 87.9 & \textbf{88.4} & \textbf{87.1} & \textbf{93.0} & \textbf{92.0} & \textbf{60.7} \\
OpenDesign & 95.2 & 69.6 & 55.0 & 75.2 & \textbf{88.3} & 85.1 & 82.3 & 80.6 & 85.5 & 21.8 \\
JarvisHub & 87.6 & 68.0 & 49.6 & 68.5 & 65.4 & 80.9 & 80.2 & 84.7 & 84.2 & 49.1 \\
\midrule
\modelgroup{\tabgroup{(b) GPT-5.6 Sol High}} \\
\ours \oursname{\system} & 97.3 & \textbf{92.5} & \textbf{79.3} & \textbf{90.2} & 87.7 & 88.8 & \textbf{88.5} & 93.8 & \textbf{92.5} & \textbf{74.0} \\
Codex & \textbf{98.1} & 78.1 & 71.3 & 75.3 & \textbf{89.8} & \textbf{88.9} & 87.9 & \textbf{94.3} & 90.8 & 44.4 \\
\bottomrule
\end{tabularx}
\end{papertable}

\paragraph{Overall comparison.}
Across 228 tasks in 19 domains (Table~\ref{tab:main-results}), \system\ leads each LLM group on constraint satisfaction (85.0\% on DeepSeek; 92.5\% on GPT), reference fidelity, and identity consistency. OpenDesign and Codex score higher on set consistency; Codex also has higher point estimates for delivery, technical quality, and text rendering. JarvisHub leaves 12.4\% of artifacts undelivered. \system's strengths thus lie in satisfying constraints and preserving object identity across outputs, one aspect of carrying design decisions forward. Appendix~\ref{app:reference-coverage} examines reference reuse and its association with identity consistency within JarvisHub.

\paragraph{Cross-domain results.}
Figure~\ref{fig:domains} tests how widely this advantage extends across the 19 domains, with 12 tasks per domain. \system\ leads or ties in 18 under each LLM. JarvisHub leads in product design on DeepSeek, and Codex in exhibition design on GPT. This broad coverage supports \emph{Anything}: our method fulfills constraints across diverse design domains.

\begin{paperfigure}[!htbp]
\centering
\includegraphics[width=\linewidth]{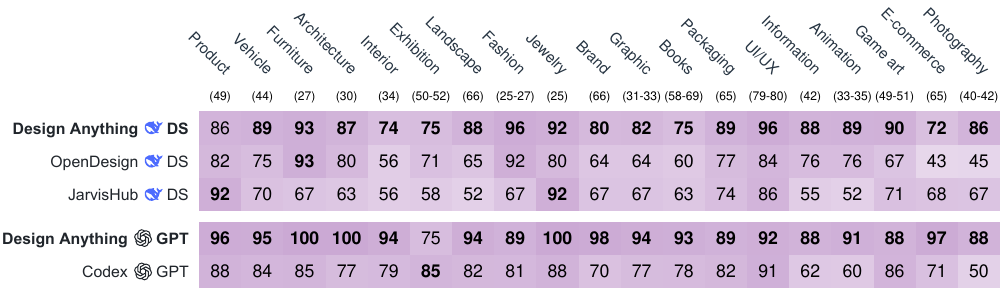}
\caption{\textbf{\system\ leads or ties in 18 of 19 domains per model.} Cells show constraint satisfaction (\%) on 12 tasks per domain; bold marks the within-model best. Parentheses give applicable-constraint counts. DS and GPT denote DeepSeek V4.1 Flash High and GPT-5.6 Sol High.}
\label{fig:domains}
\label{tab:domains}
\end{paperfigure}

\paragraph{Qualitative comparison.}
Figure~\ref{fig:comparison} compares five configurations on one 14-artifact animation brief, a railway story in which two otters swap matching cases. Both \system\ configurations and Codex are credited with all 14 artifacts, JarvisHub with 13, and OpenDesign with 11. Credit, however, does not ensure correct content: JarvisHub turns the otters into humans in its color script, Codex reuses one image for two action phases, and \system\ (DeepSeek) leaves the cases unreturned in its final keyframe. File counts alone thus cannot show whether a long-horizon project stays faithful to its brief. Enlarged previews appear in Figures~\ref{fig:compare-app-1}--\ref{fig:compare-app-2}.

\begin{paperfigure}[!htbp]
\centering
\includegraphics[width=\linewidth]{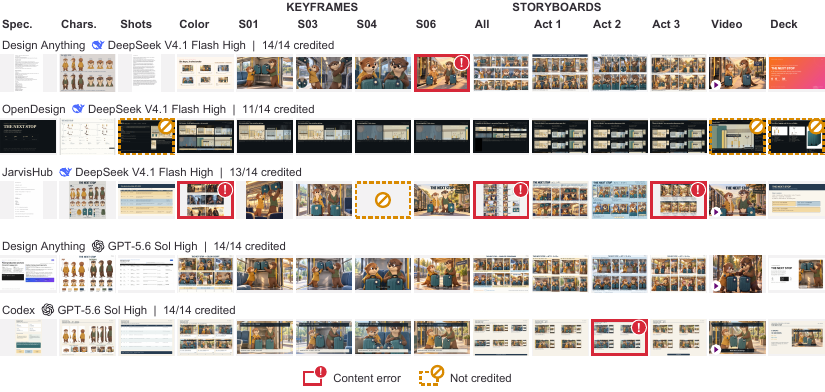}
\caption{\textbf{Same-task comparison across five configurations.} Red markers highlight selected content errors; slashed circles indicate items not credited by the evaluator.}
\label{fig:comparison}
\end{paperfigure}

\subsubsection{Results at scale}
\label{sec:analysis}
\label{sec:scale}

To test \emph{At Scale}, Figure~\ref{fig:scale} compares five configurations on 38 matched project families at 6, 14, and 24 artifacts. At 24, \system\ achieves 85.7\% constraint satisfaction on DeepSeek and 91.9\% on GPT, leading OpenDesign by 25.1\% and Codex by 23.0\%. The constraint-satisfaction advantage thus persists as the same projects require more artifacts.

\begin{paperfigure}[!htbp]
\centering
\includegraphics[width=\linewidth]{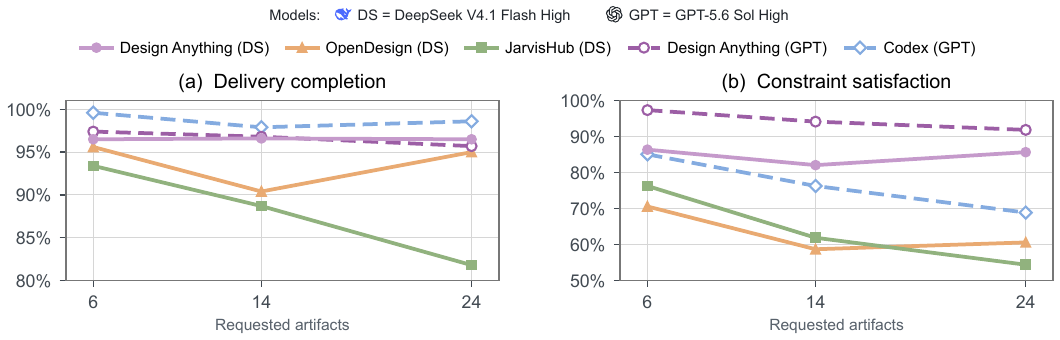}
\caption{\textbf{\system\ retains its constraint-satisfaction advantage from 6 to 24 artifacts.} Panels show (a) delivery completion and (b) constraint satisfaction on 38 matched families. Solid lines with filled markers denote DS; dashed lines with hollow markers denote GPT.}
\label{fig:scale}
\end{paperfigure}

\system's advantage at larger scales lies in satisfying the brief's constraints throughout the delivered set. On 24-artifact tasks, Codex delivers over 97\% of requested artifacts but satisfies 68.9\% of constraints, compared with 91.9\% for \system\ using the same LLM. Each constraint counts as satisfied only when it holds in every eligible artifact, so this gap reflects more reliable fulfillment across the set, beyond delivering the requested files. Paired bootstrap comparisons and sensitivity checks for missing deliveries support this conclusion (Appendix~\ref{app:robustness}).

Figure~\ref{fig:architecture-scales} illustrates how the outputs expand across three matched architectural design tasks, from spatial views and drawings to shading details, documents, and video.

\begin{paperfigure}[!htbp]
\centering
\includegraphics[width=\linewidth]{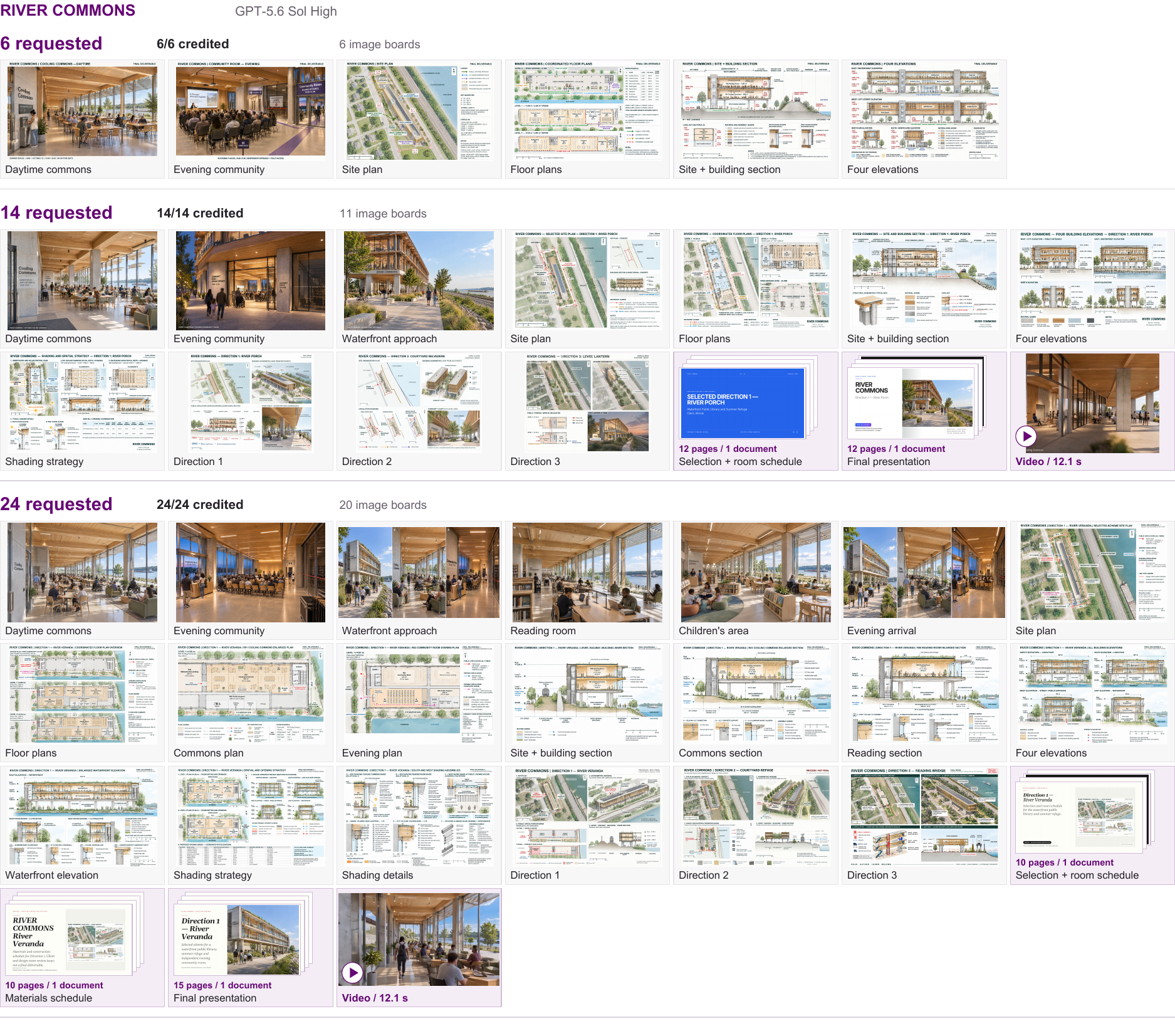}
\caption{\textbf{\system\ (GPT-5.6 Sol High) on RIVER COMMONS with 6, 14, and 24 requested artifacts.} Stacks denote documents; play icons denote videos. Counts give credited artifacts.}
\label{fig:architecture-scales}
\end{paperfigure}
\FloatBarrier

\subsection{Ablation studies}
\label{sec:execution-analysis}
\label{sec:components}

To test the task-tracking and delivery mechanisms in Figure~\ref{fig:architecture}, we compare three configurations on 19 DeepSeek V4.1 Flash High tasks requesting 24 artifacts, one per domain. The reduced harness removes the delivery loop (\figstep{185}), task contract $C$ (\figstep{183}), and structured planning tools used by the agent loop (\figstep{184}); Appendix Table~\ref{tab:continuation-config} gives the full configurations. \emph{Stop at model end} ends the run when the LLM stops calling tools; \emph{Generic prompt} instead asks it to keep working. \emph{Complete} uses the full harness.

\begin{papertable}[!htbp]
\centering\tabfont
\setlength{\tabcolsep}{6pt}
\papertableformat{continuation}
\newcommand{\gain}[1]{\,{\color{black!55}\papertablegainfont(+#1)}}
\caption{\textbf{Delivery-loop study on 19 paired 24-artifact tasks (DeepSeek V4.1 Flash High).} Parentheses give gains over Stop at model end; ``Runs $<50\%$'' counts runs below half delivery. Requests and costs are per-run means; $^\dagger$cost uses LLM list prices on 17 paired tasks with complete telemetry.}
\label{tab:continuation}
\begin{tabular}{lrrrrr}
\toprule
{\tabhead Configuration} & {\tabhead Delivery (\%)} & {\tabhead Runs $<50\%$} & {\tabhead Constraint (\%)} & {\tabhead Gen.\ requests} & {\tabhead Cost (\$)$^\dagger$} \\
\midrule
Stop at model end & 73.9 & 6 & 57.5 & 18.6 & 0.48 \\
Generic prompt & 96.9\gain{23.0} & 0 & 75.3\gain{17.8} & 36.1 & 1.30 \\
\rowcolor{oursrow} \oursname{\papertablecell{Complete (\system)}{\shortstack[l]{Complete\\(\system)}}} & 96.1\gain{22.1} & 0 & \textbf{90.4}\gain{32.9} & 24.1 & 0.43 \\
\bottomrule
\end{tabular}
\end{papertable}

Complete raises delivery from 73.9\% to 96.1\% and constraint satisfaction from 57.5\% to 90.4\% relative to Stop at model end, eliminating all six runs below half delivery (Table~\ref{tab:continuation}). Generic prompt also recovers these early stops, but issues 36.1 generation requests per run versus 24.1 for Complete. With numerically similar delivery (96.1\% versus 96.9\%), Complete incurs about one-third the mean LLM cost: \$0.43 versus \$1.30 on 17 paired tasks with complete telemetry.

Together, these mechanisms support high delivery with less repeated generation. Appendix~\ref{app:ablations} reports paired tests and further planning-interface and state-summary studies.

\FloatBarrier
\subsection{Human evaluation}
\label{sec:human}

To assess the delivered sets from a professional design perspective, twelve designers in five disciplines compare outputs on 185 tasks, yielding 1{,}019 blind pairwise judgments. They choose which set they would hand to the client, with ties allowed. \system\ is preferred over each baseline using the same LLM: designers choose it in 82\% of judgments against OpenDesign, 70\% against Codex, and 51\% against JarvisHub, compared with 37\% for JarvisHub and 12\% ties (Figure~\ref{fig:human}b).

\begin{paperfigure}[!htbp]
\centering
\includegraphics[width=0.95\linewidth]{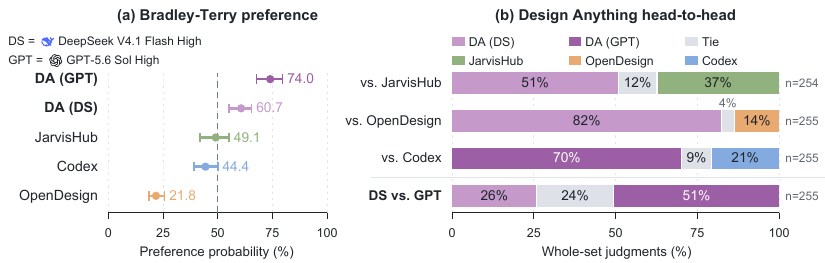}
\caption{\textbf{The two \system\ configurations have the highest estimated designer preference.} (a) Bradley--Terry estimates with 95\% task-and-designer bootstrap intervals (1{,}019 judgments, 185 tasks, 12 designers). (b) Pairwise choices with ties; $n$ counts judgments. DA denotes \system; the bold last row compares its DeepSeek and GPT configurations.}
\label{fig:human}
\end{paperfigure}

Our GPT and DeepSeek configurations rank first and second in estimated designer preference (74.0\% and 60.7\%), both with 95\% intervals above 50\% (Figure~\ref{fig:human}a). These Bradley--Terry estimates average each configuration's fitted preference probabilities against the other four; direct comparisons between our two configurations connect the LLM groups. The pairwise choices and joint ranking show that designers more often select \system's delivered sets for client handoff, complementing the automatic measures with professional judgment of the project as a whole. Appendix~\ref{app:designer-evaluation} gives the evaluation and estimation details.
\FloatBarrier

\paperbeforeconclusion
\section{Conclusion}
\label{sec:conclusion}
We presented \system, a harness that keeps design intent outside the model in an intent--execution graph, which binds each deliverable to the workflow steps that produce it, and \benchmark, 228 long-horizon design tasks across 19 domains and three project scales. With the LLM fixed, \system\ satisfies more of the brief's constraints than the strongest baseline overall and at every project scale, leads or ties in 18 of 19 domains, and receives the highest designer preference. In the study of the delivery loop (Section~\ref{sec:components}), the full harness delivers nearly as many artifacts as a harness that answers every stop with a generic prompt to keep working, at about one-third of its LLM cost. Limitations are discussed in Appendix~\ref{app:limitations}.

\label{main-text-end}

\ifpaperstatements
\input{sections/statements}
\fi
{
\paperbibliographyfont
\bibliography{references}
\bibliographystyle{\paperbibliographystyle}
}

\clearpage
\paperappendixsetup
\appendix
\section{Implementation and Runtime Details}
\label{app:method-details}

\subsection{Task contract}

The task contract, the intent layer of the intent--execution graph, is stored as a structured project object visible on the canvas. It holds the final deliverable description, stable deliverable identifiers, public success criteria, constraints, decisions marked as coming from the user or the agent, and open questions. It holds no execution phase or completion counter, so revising it updates intent rather than recording progress.

\subsection{Graph construction and artifact ownership}

The binding $\beta$ is implemented through delivery groups, named sets of nodes. Graph-building operations accept typed node descriptions, configurations, connections, and the delivery group that receives them. They can create new nodes or associate eligible existing nodes with that group. Groups explicitly declare which contract deliverables they implement; an empty declaration denotes supporting work. Shared groups require output ownership information so that an artifact for one deliverable is not automatically counted toward another. Dependency declarations and port-level edges serve different purposes: the former express ordering between groups, while the latter carry concrete inputs to operations.

\subsection{Operational delivery checks}

For each deliverable, the controller finds its bound nodes through the delivery groups that claim it and through node ownership, and examines the status of the bound executable nodes; data-only inputs and annotations do not count as artifact-generating execution. In groups shared by several deliverables, an execution failure that cannot be attributed to one item remains a group-level failure signal rather than being assigned to an item.

Output-form checks use node capability declarations. An image-producing operation cannot satisfy a readable-document deliverable merely because its image resembles a document, and a minimum-page declaration requires an implementation capable of paginated output without verifying the realized page count. Quantity checks count supported output items rather than generation nodes. The benchmark's artifact evaluation is therefore necessary even when every bound node has finished.

\subsection{Observation and new turns}

The state summary is a compact projection of workflow state regenerated between turns; detailed read tools expose individual nodes, connections, configurations, and outputs.

After a turn ends, the controller reads the workflow graph and the execution records and lists obligations for the root goal, deliverables, decisions, required questions, and execution. Each obligation is open, waiting, in progress, satisfied, failed, or cancelled. The controller then selects the next action in this order:
\begin{enumerate}
    \item Stop on cancellation, loss of edit access, execution failures, or a repeated-error limit.
    \item Wait if a turn, job, or delegated execution is still active, or if a decision requested from the user is still pending.
    \item Terminate unsuccessfully on failed obligations, or successfully when the root obligation is satisfied and no unresolved obligation remains. The root obligation is derived, not declared: it is satisfied only when every deliverable is delivered and no required question, pending decision, or running execution remains.
    \item For remaining work, check progress and resource limits; either start another model turn or stop with an explicit reason.
\end{enumerate}

The message starting a new turn names the outstanding deliverables and, for each, the blocking reason, such as a missing output form or a quantity shortfall. The model then decides whether to read state, modify a contract, build or edit a graph, execute an operation, delegate work, or request a decision.

Progress monitoring separates obligation changes from observable workflow changes. Repeatedly restating a plan is not sufficient evidence of artifact progress. Progress limits stop repeated new turns that make no progress, while allowing active execution and pending decisions; the thresholds and run limits are listed in Appendix~\ref{app:run-configuration}.

\subsection{Automatic verifier}

When every deliverable is delivered, an automatic verifier reviews the workflow graph in the background: deterministic checks first list structural problems, such as unconnected nodes, and the verifier checks them against the success criteria and returns concrete findings to the agent. Visual design critique is available only as a tool the agent may call. Goal completion remains governed by the control layer. The verifier is enabled in the main comparison and the Complete delivery-loop configuration, and disabled in the two reduced delivery-loop configurations and all planning configurations. Recorded run settings are in Appendix~\ref{app:run-configuration}; component-study configurations are in Appendix~\ref{app:ablations}.

\section{Benchmark Construction and Evaluation Protocol}
\label{app:benchmark}
This appendix describes task construction, output processing, metric definitions, judge calibration, and the designer study. Complete results and sensitivity analyses are in Appendix~\ref{app:systems}; controlled component studies and visual evidence follow in Appendices~\ref{app:ablations} and~\ref{app:cases}.

\subsection{Task construction and benchmark independence}
\label{app:benchmark-protocol-details}

The 19 domains are product design; vehicle styling; furniture and home products; architecture; interior design; exhibition design; landscape design; fashion and textiles; jewelry and accessories; brand identity; graphic design and advertising; books and illustration; packaging; UI/UX; information design and data visualization; animation and storyboarding; game art and worldbuilding; e-commerce product imagery; and advertising and fashion photography. Figure~\ref{fig:benchmark-overview} summarizes their equal representation. The task counts below distinguish scale variants from distinct project identities.

\begin{center}
\tabtextfont
\setlength{\tabcolsep}{4pt}
\begin{tabularx}{\linewidth}{@{}Yr*{3}{>{\centering\arraybackslash}p{0.06\linewidth}}rr@{}}
\toprule
{\tabhead Task condition} & {\tabhead Tasks/domain} & \multicolumn{3}{c}{{\tabhead Tasks by artifact count}} & {\tabhead Total tasks} & {\tabhead Projects} \\
\cmidrule(lr){3-5}
 & & {\tabhead 6} & {\tabhead 14} & {\tabhead 24} & & \\
\midrule
Matched scales & $2\times3=6$ & 38 & 38 & 38 & 114 & 38 \\
Reference materials & 3 & 26 & 31 & 0 & 57 & 57 \\
Breadth & 3 & 19 & 38 & 0 & 57 & 57 \\
\midrule
Total & 12 & 83 & 107 & 38 & 228 & 152 \\
\bottomrule
\end{tabularx}
\end{center}

Each domain contributes two base project families: one asks the system to propose alternatives and select one, and the other fixes the direction in the brief. Each family has three scale versions with shared background and inputs, yielding $19\times2\times3=114$ tasks. Only the 14- and 24-artifact versions request process artifacts, such as alternative directions and specifications. The three reference-material and three breadth projects per domain are independent projects, adding 114 tasks. Each requests either 6 or 14 artifacts, with no additional scale variants of the same project. Thus $19\times(2\times3+3+3)=228$ tasks correspond to $19\times(2+3+3)=152$ project identities. Only the base tasks support within-family scale comparisons.

\paragraph{Evaluation specifications.}
Each specification translates the public brief and supplied materials into scoring criteria. Requested categories specify expected quantities, mark core categories, and distinguish process from final deliverables. Additional rules specify factual constraints, source and target artifacts for comparisons, and when a criterion is inapplicable. Systems receive only the English brief and materials.

\paragraph{Benchmark independence.}
\label{app:benchmark-independence}
Tasks were authored with LLM assistance from a per-domain topic plan and reviewed by the authors, with English scenarios as the only system input. No internal label of \system{} enters scoring. During development, \system{} was run on a separate 20-task pilot corpus and on small subsets of the final corpus (5-task subsets and 13 base-scale tasks) to fix infrastructure failures. No brief was revised in response to any system's score except under a recorded defect rule applied identically to all systems. Each system kept whatever general-purpose skills its product bundles; none used skills written for a particular design domain, which could supply prebuilt design procedures, and none received task-specific hints.

Execution and scoring use frozen task specifications, and each run is scored against the specification of the brief it received. The three tasks of one book project family (6, 14, and 24 artifacts) exist in two versions: OpenDesign and JarvisHub received the earlier archival photography brief, while both \system{} configurations and Codex received the later historical astronomy plate brief; the other 225 tasks have identical briefs and specifications for all systems.

\subsection{Output processing and assignment}
\label{app:output-interface}

The evaluator consumes a package of delivered files and evaluation specifications, independent of the system's internal representation. It can therefore compare a canvas-based workflow with a file-based agent without requiring the latter to reconstruct a canvas. Import records distinguish supplied inputs, primary deliverables, and derived representations. Multiple file formats of the same work are associated to avoid counting a preview and its source as two deliveries. Input materials copied into an output directory remain inputs, identified by their content.

HTML decks are rendered by their own page structure, and PDF and presentation files keep their page and text information; a file whose pages cannot be read makes a page minimum unmeasurable rather than failed. Rendered deck pages can be inspected for consistency but do not count as separately requested artifacts.

To assign images to requested categories, the classifier scores each category and an unassigned option for every image; the assignment fills each category up to its expected quantity and uses each image at most once. Remaining images may be recorded as overflow without raising completion above the expected quantity. This matching uses the artifact and the specification, not \system{} node labels, group names, or execution order. The classifier and evaluator versions are given in Appendix~\ref{app:systems}, and the classifier's calibration in Appendix~\ref{app:judge-calibration}.

\subsection{Measures, eligibility, and aggregation}
\label{app:measures}
\label{app:eligibility}
Table~\ref{tab:measures} summarizes the measures used in the main comparison. The definitions below follow the same order and specify the evidence and counting unit of each measure.

\begin{papertable}[h]
\centering\tabtextfont
\setlength{\tabcolsep}{3.5pt}
\caption{\textbf{Measures of \benchmark.} The measures assess fulfillment of the brief, agreement with the client's materials or the run's own outputs, design quality, and designer preference between systems on the same brief.}
\label{tab:measures}
\begin{tabularx}{\linewidth}{@{}l>{\raggedright\arraybackslash}p{0.2\linewidth}Yl@{}}
\toprule
{\tabhead Measure} & {\tabhead Compared against} & {\tabhead Definition} & {\tabhead Scored by} \\
\midrule
\multicolumn{4}{@{}l}{\tabgroup{Requirement fulfillment}} \\
Delivery completion & Brief: categories, quantities & Requested files delivered, capped per category & Classifier, count \\
Constraint satisfaction & Brief: factual constraints & Factual constraints satisfied in every targeted artifact (Eq.~\ref{eq:dc-cs}) & Criterion check \\
Reference fidelity & Client's reference assets & Output--reference pairs depicting the same object & Calibrated judge \\
\addlinespace
\multicolumn{4}{@{}l}{\tabgroup{Cross-artifact consistency}} \\
Identity consistency & The run's anchor artifact & Output--anchor pairs depicting the same object & Calibrated judge \\
Set consistency & The run's whole set & 100-point rubric score for shared visual rules & Calibrated judge \\
\addlinespace
\multicolumn{4}{@{}l}{\tabgroup{Design quality}} \\
Technical, aesthetics, text, domain & Image checklist and delivery requirements & Checklist mean $\times$ delivery completion (100-point scale)~\citep{li2026qwenimagebench} & VLM judge, count \\
\addlinespace
\multicolumn{4}{@{}l}{\tabgroup{Human evaluation}} \\
Preference & Another system, same brief & Blinded pairwise choice; Bradley--Terry strength & Designers \\
\bottomrule
\end{tabularx}
\end{papertable}

\paragraph{Delivery completion.}
For requested category $i$ with expected and delivered quantities $e_i$ and $d_i$, delivery completion is $\mathrm{DC}=\sum_i \min(d_i,e_i)/\sum_i e_i$; excess outputs in one category cannot offset a shortfall in another. A file's category assignment and delivered status are distinct from its compliance with format or content requirements. Page minima belong to the supplementary task-level outcomes in Appendix~\ref{app:task-outcomes}, rather than delivery completion.

\paragraph{Constraint satisfaction.}
Constraint satisfaction applies each constraint-specific criterion to its eligible targets. Let $K$ be the set of applicable factual constraints, $T(c)$ the target set for constraint $c$, and $q_c(y)\in\{0,1\}$ indicate whether artifact $y$ satisfies it. With the target eligibility and missing-delivery rules below, the task-level score is
\begin{equation}
\mathrm{CS}(Y;x)=\frac{1}{|K|}\sum_{c\in K}\mathbf{1}\!\left[\,q_c(y)=1\ \ \forall\, y\in T(c)\right].
\label{eq:dc-cs}
\end{equation}
Most targets are generated images (85--89\% of target files for every system).

\paragraph{Constraint targets and unavailable evidence.}
\label{app:constraint-eligibility}
Targets are delivered files in the named categories, capped at the requested quantity, or only each category's final file when the specification says so. The following rules distinguish absent deliveries from unavailable evidence:
\begin{itemize}
\item If none of the target categories contains a delivered file, the constraint counts as unsatisfied and stays in the denominator. If only some categories are empty, the constraint is checked on the delivered files alone.
\item Files whose text transcription failed are removed from the target set. The constraint is excluded if no targeted file remains after this removal.
\item The constraint is also excluded when its target categories are empty but a delivered file of an accepted type received no valid category from the classifier.
\end{itemize}
\paragraph{Constraint check.}
For each constraint, the evaluator transcribes artifact text, removes whitespace and dash characters, and normalizes full-width forms before exact, case-sensitive matching against the required wording. This defines $q_c(y)$.

\paragraph{Reference fidelity and cross-artifact consistency.}
Consistency comparisons are scoped to properties that the project is supposed to preserve. Alternative concepts need not look identical, and a legitimate refinement need not reproduce a sketch literally. The reference source determines the measure: agreement with a supplied reference asset is reference fidelity, a requirement-fulfillment measure, whereas agreement among outputs of the same run is identity consistency or set consistency. Unavailable pair judgments are distinguished from mismatches: those caused by the system's own outputs stay in the denominators of reference fidelity and identity consistency, and those caused by an external reference are excluded.

\paragraph{Set consistency.}
The reported set-consistency score uses only rubric items labeled consistency, excluding separate compliance items. Item scores of 0, 1, and 2 map to 0, 60, and 100; inapplicable items are excluded and unmet items contribute zero. Within a task, the evaluator averages rule-level consistency means equally. Across tasks, it weights these task scores by their numbers of scored consistency items.

\paragraph{Design quality.}
The image-checklist component of design quality covers technical quality, aesthetics, text rendering, and domain-specific criteria. A criterion may be inapplicable, for example anatomical fidelity in an image without a person. The scores in Appendix~\ref{app:project-design-quality} use a checklist rubric judged by the same DeepSeek Flash model used for category assignment (Appendix~\ref{app:systems}), with reasoning disabled. The judge sees each delivered image of a core image category on its own and marks every checklist item 0, 1, 2, or not applicable. Item marks are averaged within each sub-criterion, the sub-criteria are averaged within each of the four groups, and the result is mapped to a 100-point scale. A task's score for a group averages the images within each core category and then averages over categories, so a category that asks for many images does not outweigh the others.

Let $Q_{t,g}$ be the image-checklist score for task $t$ and group $g$. A task has no score for a group when none of its judged images has an applicable item in that group; such a task is left out of that group's average rather than counted as zero. With $\mathcal{T}_g$ the set of tasks that have a score for group $g$, design quality is
\begin{equation}
\mathrm{DQ}_{g}=\mathrm{DC}\cdot\frac{1}{|\mathcal{T}_g|}\sum_{t\in\mathcal{T}_g}Q_{t,g},
\label{eq:project-design-quality}
\end{equation}
where $\mathrm{DC}\in[0,1]$ is the system's pooled delivery completion over all requested artifact types, as defined above. Both factors are kept at full precision, and only the product is rounded. Design quality combines the visual quality of delivered images with project delivery coverage; missing work reduces all four dimensions, and excess files do not offset shortfalls in other categories. It aggregates image-level assessments rather than introducing a whole-set visual rubric. Constraint satisfaction and reference fidelity separately assess compliance with the brief and supplied assets; set consistency assesses shared visual rules.

\paragraph{Rendering and missing scores.}
On pages that an importer rendered from HTML or rasterized from vector files, the three items for noise, edge clarity, and resolution are marked not applicable, because they measure our renderer rather than the design. An image whose judge call fails is left unscored. This judge has not been calibrated against designers on the present artifacts; Appendix~\ref{app:project-design-quality} instead reports a second judge on a stratified sample.

\paragraph{Aggregation.}
Delivery completion, constraint satisfaction, reference fidelity, and identity consistency pool their counting units over tasks. In particular, the reported constraint-satisfaction rate pools passed and applicable constraints; Equation~\ref{eq:objective} instead expresses expected task-level constraint satisfaction. Set consistency averages rule-level scores within a task and weights tasks by their numbers of scored items. For design quality, image-checklist scores average within each task and then over tasks, and the resulting mean is multiplied by pooled delivery completion (Equation~\ref{eq:project-design-quality}). Appendix~\ref{app:metric-counts} reports the corresponding observed counts.

\subsection{Judge calibration}
\label{app:judge-calibration}

For the identity and set-consistency checks below, two professional designers independently labeled samples drawn from the final benchmark artifacts of the three DeepSeek-backed systems. The first five items of each sample are warm-up items and do not enter the statistics. A judge qualifies when Cohen's $\kappa\ge0.60$ and raw agreement is at least 85\%. Human-judge agreement is computed on the items where both designers agree.

\paragraph{Identity judge.} A single judge decides whether two images depict the same object, character, logo, or space; it serves both reference fidelity and identity consistency. Of 108 sampled pairs, 103 are analyzed. Human-human agreement is $\kappa=0.931$ with 96.1\% raw agreement. On the 99 consensus pairs, human-judge agreement is $\kappa=0.887$ with 93.9\%. By reference source, $\kappa$ is 0.932 for pairs within a run (identity consistency, $n=56$) and 0.830 for pairs against supplied assets (reference fidelity, $n=43$). By subject, $\kappa$ is 1.000 for characters, 1.000 for logos, 0.805 for products, and 0.826 for spaces.

\paragraph{Set-consistency judge.} Of 117 sampled rubric items labeled consistency, 112 are analyzed. Human-human agreement is $\kappa=0.893$ with 92.9\%. On the 104 consensus items, human-judge agreement is $\kappa=0.885$ with 92.3\%. Removing the items that the judge answered not applicable leaves 100 analyzed items. On them, human-judge $\kappa$ is 0.804 and 0.836 for the two designers and human-human $\kappa$ is 0.901, so the qualification does not rest on not-applicable answers.

\paragraph{Direction of judge error.} Agreement statistics do not show whether judge errors favor a particular system, so we also compare the judge with the designers per system on the consensus items. The identity judge labels more pairs as the same subject than the designers do: by 2.9 points for \system{}, 2.9 for OpenDesign, and 14.3 for JarvisHub. On the 100-point scale, the set-consistency judge scores \system{} 9.2 points below the designers, while it scores JarvisHub 1.6 and OpenDesign 2.0 points above them. Neither judge's error favors \system{}.

\paragraph{Category assignment.}
The classifier assigns delivered artifacts to the requested categories used to compute delivery completion. Calibration covers 120 tasks. Human-human agreement is Cohen's $\kappa=0.89$; human-classifier agreement is $\kappa=0.83$, using the two annotators' consensus labels as the reference.

\paragraph{Text transcription.}
For text transcription used in constraint checks, human-human agreement is Cohen's $\kappa=0.98$. Human-machine agreement is $\kappa=0.92$ on the subset where the two annotators agree, using their consensus annotations as the reference.

\subsection{Designer evaluation protocol}
\label{app:designer-evaluation}
\paragraph{Participants and assignment.}
The designer study (Section~\ref{sec:human}) uses the same run of each system that the automatic measures score, so no run is selected by appearance. Twelve designers rate 185 tasks in five discipline groups: visual communication (four designers; brand identity, graphic design and advertising, packaging, books and illustration, information design, photography, and e-commerce imagery), digital interaction (two; UI/UX, game art, animation and storyboarding), product (two; product design, vehicle styling, furniture), spatial (two; architecture, interior, exhibition, and landscape design), and fashion and accessories (two). Within a group, tasks are stratified by domain, task condition, and project scale and dealt to designers in rotation; each designer first rates three warm-up tasks, and some tasks are rated a second time by another designer of the group.

\paragraph{Pairings and displayed evidence.}
Baseline comparisons pair systems within an LLM: \system, OpenDesign, and JarvisHub pairwise on DeepSeek, and \system\ against Codex on GPT; \system\ on DeepSeek is also compared directly with \system\ on GPT, which connects the two model groups on one Bradley--Terry scale. Each comparison shows every file the evaluator assigned to a requested category, in the brief's deliverable order.

\paragraph{Whole-set judgments and estimation.}
Designers see the brief with a reference translation, never system names, reasoning, or completion claims. Sides are randomized. They choose the set they would rather hand to the client, answering left, right, or approximately equal. The study yielded 1{,}019 whole-set judgments, which support the results in Section~\ref{sec:human}. Whole-set preference is estimated with a Bradley--Terry model~\citep{bradley1952rank}, $\Pr(i\succ j)=e^{\beta_i}/(e^{\beta_i}+e^{\beta_j})$, fitted to the 1{,}019 judgments. A tie contributes half a win to each side. For each of the five configurations, we average its fitted preference probabilities against the other four, weighting opponents equally regardless of the number of observed comparisons. This is the probability of beating a uniformly sampled other configuration, not a raw head-to-head win fraction. We report 95\% intervals from 2{,}000 bootstrap replicates over tasks and designers.

\section{System Configurations and Complete Results}
\label{app:systems}
\subsection{Controlled comparisons and experimental settings}
Table~\ref{tab:tracks} identifies what each reported comparison holds fixed and what it varies. The two main model groups compare complete systems: holding the LLM fixed does not hold media-generation tools, orchestration, or execution policies fixed, so their differences describe complete system configurations. No Codex run reached its 150-minute limit.

\begin{papertable}[h]
\centering\tabtextfont
\caption{\textbf{Controlled comparison matrix.} The final column states the comparison of interest; the rows include both interventions and observational analyses. The main model groups share briefs except on three tasks of one book project family (Appendix~\ref{app:benchmark-protocol-details}) and use a common evaluator. The delivery-loop study uses the same briefs and evaluator across its configurations; the planning and state-summary studies use separate tasks and harness versions (Appendix~\ref{app:ablations}).}
\label{tab:tracks}
\label{tab:systems}
\begin{tabularx}{\linewidth}{@{}>{\raggedright\arraybackslash}p{0.2\linewidth}rY>{\raggedright\arraybackslash}p{0.2\linewidth}@{}}
\toprule
{\tabhead Comparison} & {\tabhead Tasks} & {\tabhead Held fixed} & {\tabhead Varied factor} \\
\midrule
DeepSeek group (Table~\ref{tab:main-results}a) & 228 & DeepSeek V4.1 Flash High for every agent and sub-agent; gpt-image-2.5-sunburst image generation & Harness and agent: \system, OpenDesign, JarvisHub \\
GPT group (Table~\ref{tab:main-results}b) & 228 & GPT-5.6 Sol High; gpt-image-2.5-sunburst image generation & Harness and agent: \system, Codex \\
Project scale (Section~\ref{sec:scale}) & $38\times3$ & Project family, background, and inputs; system & Requested artifacts: 6, 14, 24; process artifacts \\
Reference coverage (Appendix~\ref{app:reference-coverage}) & 228 & DeepSeek V4.1 Flash High; within JarvisHub, the harness & Reference coverage (observed, not assigned) \\
Delivery loop (Section~\ref{sec:components}) & 19 & \system\ on DeepSeek; 24-artifact tasks, one per domain & Delivery loop and other channels for tracking the task \\
Planning interface (Appendix~\ref{app:planning-study}) & 8 & \system; other channels for tracking the task & Planning tools: structured, text, none \\
State summary (Appendix~\ref{app:digest-study}) & 10 pairs & \system; direct read tools & Summary on or off \\
Human evaluation (Section~\ref{sec:human}) & 185 & Scored runs of each system & System within a model; the two \system\ configurations compared directly \\
\bottomrule
\end{tabularx}
\end{papertable}

All five configurations use the same 228 task identifiers, with frozen evaluation specifications matched to the brief version each received (corpus v0.6; Appendix~\ref{app:benchmark-protocol-details}). Four configurations are scored with evaluator v1.18. OpenDesign is scored with evaluator v1.19, which reads HTML deliverables as a browser displays them: it runs the local scripts they reference, and a full-page rendering may fill an image slot when the HTML file itself fills none. Only OpenDesign's HTML deliverables draw their content with scripts, so only its runs were re-imported and rescored. Category assignment uses \texttt{deepseek-flash} with reasoning disabled, the same model alias used for the image-checklist component of design quality.

Each task contributes one final result per configuration. The reported tables aggregate these per-task evaluation records. Independent scorings of the same outputs differ by about three points on judged measures, so we do not interpret differences of that size.

\paragraph{Recorded runtime configuration.}
\label{app:run-configuration}
Table~\ref{tab:run-config} lists the recorded settings; Appendix~\ref{app:method-details} describes the mechanisms.

\begingroup
\tabtextfont
\setlength{\LTcapwidth}{\linewidth}
\begin{longtable}{@{}p{0.24\linewidth}p{\dimexpr0.76\linewidth-2\tabcolsep\relax}@{}}
\caption{\textbf{Recorded runtime configuration of \system.} Settings apply to both configurations unless noted.}
\label{tab:run-config}\\
\toprule
{\tabhead Setting} & {\tabhead Recorded configuration} \\
\midrule
\endfirsthead
\toprule
{\tabhead Setting} & {\tabhead Recorded configuration (continued)} \\
\midrule
\endhead
\bottomrule
\endfoot
Harness version & DeepSeek V4.1 Flash High runs: one controller version. GPT-5.6 Sol High runs: a later version of the same code that adds the retries listed under Controller limits and updates image analysis and image repair. \\
Agent and role models & Every role (orchestrator, builder, verifier, summarizer, and specialist helpers), text-generation nodes, and image analysis use the configuration's LLM. High in the configuration names denotes the reasoning setting: DeepSeek V4.1 Flash runs at its default high reasoning, with reasoning disabled for image analysis; GPT-5.6 Sol receives reasoning effort high for the main agent and specialist helpers. \\
Generation backends & Images: gpt-image-2.5-sunburst, medium quality, 2K. Video: a third-party video model. \\
Skills & General-purpose skills bundled with the product; none written for a particular design domain \\
Other tools & Workspace reading and editing, contract and workflow construction, synchronous and batched execution, on-demand image analysis, clarifying questions, web search, memory, and delegation to helper agents \\
Context policy & Tool results are elided beyond 50\% of the context window (1M tokens for DeepSeek V4.1 Flash High, 272k for GPT-5.6 Sol High); history is summarized beyond 85\%, keeping the last six steps; the state summary is always on \\
Resource limits & Per turn: 120 tool calls, 120 model steps, 90 minutes. Per run: 100 turns. Per task: 150 minutes wall clock (DeepSeek V4.1 Flash High) or 150--240 minutes (GPT-5.6 Sol High). Tokens are recorded but not limited. \\
Checkpoint policy & A scripted user accepts approval and review requests, picks the recommended option (otherwise the first option that does not defer), and answers free-text questions with a fixed message stating that the brief is complete; a repeated question fails the run \\
Automatic verifier & On in the main comparison and the Complete delivery-loop configuration; off in the two reduced delivery-loop configurations and all planning configurations (Appendix~\ref{app:ablations}) \\
Controller limits & The run stops after two controller decisions without obligation progress, after three without new output, or after $\min(20,\max(2,3n))$ consecutive new turns for $n$ obligations; at most 100 turns per run. GPT-5.6 Sol High runs also retry on rate limits and transport interruptions. \\
\bottomrule
\end{longtable}
\endgroup

\subsection{Metric counts and aggregate results}
\label{app:metric-counts}
Table~\ref{tab:metric-counts} gives the observed numerators, denominators, and coverage counts underlying the main results. Scoring definitions and eligibility rules are in Appendix~\ref{app:benchmark}; differences in applicable evidence explain why some denominators vary across systems.

\begin{papertable}[h]
\centering\tabfont
\setlength{\tabcolsep}{4pt}
\caption{\textbf{Counts underlying Table~\ref{tab:main-results}.} Columns count delivered/requested items (delivery completion), passed/applicable constraints (constraint satisfaction), same/eligible pairs against supplied reference assets (reference fidelity), same/eligible pairs within the run (identity consistency), and scored rubric items (set consistency). JarvisHub's reference-fidelity denominator is a range: 1,326 counts the judge's abstentions as mismatches and 1,127 excludes them. The last column is a coverage count, not a simple pooled-mean denominator.}
\label{tab:metric-counts}
\begin{tabular}{lrrrrr}
\toprule
{\tabhead System} & {\tabhead Delivery} & {\tabhead Constraints} & {\tabhead Reference} & {\tabhead Identity} & {\tabhead Set items} \\
\midrule
\rowcolor{oursrow} \oursname{\systemshort{} (DS)} & 2809/2908 & 767/902 & 1063/1407 & 554/656 & 606 \\
OpenDesign (DS) & 2768/2908 & 616/885 & 768/1396 & 467/621 & 585 \\
JarvisHub (DS) & 2547/2908 & 608/894 & 658/(1127--1326) & 411/600 & 502 \\
\midrule
\rowcolor{oursrow} \oursname{\systemshort{} (GPT)} & 2830/2908 & 834/902 & 1122/1414 & 598/663 & 614 \\
Codex (GPT) & 2853/2908 & 699/895 & 1016/1425 & 498/661 & 550 \\
\bottomrule
\end{tabular}
\end{papertable}

The primary counts follow the unavailable-evidence rules in Appendix~\ref{app:eligibility}. Table~\ref{tab:abstention} shows how excluding judge abstentions changes the two pair measures.

\FloatBarrier
\subsection{Results by task condition}
Table~\ref{tab:scale-results} checks the five configurations across three project scales, breadth tasks, and reference-material tasks. \system\ has the highest constraint satisfaction within each model group in every condition, reaching 85.3\% and 96.0\% on breadth tasks. OpenDesign delivers more on breadth tasks under DeepSeek, and Codex delivers more at every base scale under GPT. These results extend the \emph{Anything} and \emph{At Scale} findings across conditions while distinguishing brief alignment from delivery quantity.

Each base scale contains the same 38 project families, so the three scale rows of a family are dependent observations. Set consistency per condition uses the same nested aggregation as Table~\ref{tab:main-results}, applied to the tasks of that condition.

\begin{papertable}[!htbp]
\centering\tabtextfont
\setlength{\tabcolsep}{2.2pt}
\caption{\textbf{\system\ leads same-model baselines in constraint satisfaction in every task condition.} Shaded rows are \system; bold marks the best delivery, constraint, and identity score within each LLM group and condition. These scores are percentages; set consistency uses a 100-point scale. Supplementary P/F/U counts denote task passes/failures/unmeasurable outcomes for all or final deliverables (Appendix~\ref{app:task-outcomes}); they are distinct from constraint satisfaction. DS: DeepSeek V4.1 Flash High; GPT: GPT-5.6 Sol High.}
\label{tab:scale-results}
\newcolumntype{S}[1]{>{\hsize=#1\hsize\linewidth=\hsize\centering\arraybackslash}X}
\begin{tabularx}{\linewidth}{@{}lS{1.0}S{1.25}S{0.8}S{0.55}S{1.2}S{1.2}@{}}
\toprule
& \multicolumn{4}{c}{\tabhead Evaluation measures} & \multicolumn{2}{c}{\tabhead Supplementary outcomes} \\
\cmidrule(lr){2-5}\cmidrule(l){6-7}
{\tabhead System} & {\tabhead Delivery} & {\tabhead Constraints} & {\tabhead Identity} & {\tabhead Set} & {\tabhead \shortstack{P/F/U\\(all)}} & {\tabhead \shortstack{P/F/U\\(final)}} \\
\midrule
\multicolumn{7}{l}{\tabgroup{6 artifacts (38 tasks)}} \\
\rowcolor{oursrow} \oursname{\systemshort{} (DS)} & \textbf{96.5} & \textbf{86.4} & \textbf{87.7} & 93.6 & 27/10/1 & 27/10/1 \\
OpenDesign (DS) & 95.6 & 70.6 & 78.1 & 86.9 & 22/14/2 & 22/14/2 \\
JarvisHub (DS) & 93.4 & 76.3 & 76.9 & 75.0 & 21/15/2 & 21/15/2 \\
\cmidrule(lr){1-7}
\rowcolor{oursrow} \oursname{\systemshort{} (GPT)} & 97.4 & \textbf{97.4} & \textbf{97.0} & 94.9 & 31/7/0 & 31/7/0 \\
Codex (GPT) & \textbf{99.6} & 85.1 & 93.9 & 92.1 & 32/6/0 & 32/6/0 \\
\midrule
\multicolumn{7}{l}{\tabgroup{14 artifacts (38 tasks)}} \\
\rowcolor{oursrow} \oursname{\systemshort{} (DS)} & \textbf{96.6} & \textbf{82.1} & \textbf{79.0} & 86.7 & 21/15/2 & 24/12/2 \\
OpenDesign (DS) & 90.4 & 58.7 & 61.0 & 82.7 & 12/23/3 & 12/23/3 \\
JarvisHub (DS) & 88.7 & 61.9 & 71.9 & 59.4 & 7/29/2 & 12/24/2 \\
\cmidrule(lr){1-7}
\rowcolor{oursrow} \oursname{\systemshort{} (GPT)} & 96.8 & \textbf{94.2} & \textbf{89.9} & 82.7 & 20/17/1 & 25/12/1 \\
Codex (GPT) & \textbf{97.9} & 76.3 & 69.6 & 85.2 & 15/23/0 & 19/18/1 \\
\midrule
\multicolumn{7}{l}{\tabgroup{24 artifacts (38 tasks)}} \\
\rowcolor{oursrow} \oursname{\systemshort{} (DS)} & \textbf{96.5} & \textbf{85.7} & \textbf{86.3} & 87.0 & 14/24/0 & 15/23/0 \\
OpenDesign (DS) & 95.0 & 60.6 & 68.7 & 87.7 & 19/18/1 & 19/18/1 \\
JarvisHub (DS) & 81.8 & 54.4 & 60.3 & 61.0 & 4/33/1 & 7/30/1 \\
\cmidrule(lr){1-7}
\rowcolor{oursrow} \oursname{\systemshort{} (GPT)} & 95.7 & \textbf{91.9} & \textbf{87.8} & 87.2 & 14/22/2 & 15/21/2 \\
Codex (GPT) & \textbf{98.6} & 68.9 & 66.7 & 91.1 & 19/18/1 & 20/17/1 \\
\midrule
\multicolumn{7}{l}{\tabgroup{Breadth (57 tasks)}} \\
\rowcolor{oursrow} \oursname{\systemshort{} (DS)} & 96.1 & \textbf{85.3} & \textbf{85.9} & 85.0 & 30/22/5 & 32/20/5 \\
OpenDesign (DS) & \textbf{97.5} & 73.4 & 80.1 & 91.7 & 33/22/2 & 33/22/2 \\
JarvisHub (DS) & 91.6 & 74.3 & 71.8 & 69.6 & 26/28/3 & 30/24/3 \\
\cmidrule(lr){1-7}
\rowcolor{oursrow} \oursname{\systemshort{} (GPT)} & \textbf{98.8} & \textbf{96.0} & \textbf{88.8} & 88.3 & 41/15/1 & 41/15/1 \\
Codex (GPT) & 97.7 & 80.3 & 77.5 & 90.3 & 30/24/3 & 34/19/4 \\
\midrule
\multicolumn{7}{l}{\tabgroup{Ref. materials (57 tasks)}} \\
\rowcolor{oursrow} \oursname{\systemshort{} (DS)} & \textbf{97.3} & \textbf{85.5} & 81.1 & 89.7 & 31/23/3 & 31/23/3 \\
OpenDesign (DS) & 97.1 & 80.0 & \textbf{90.5} & 89.6 & 35/21/1 & 35/21/1 \\
JarvisHub (DS) & 88.8 & 70.2 & 71.4 & 63.9 & 24/32/1 & 24/32/1 \\
\cmidrule(lr){1-7}
\rowcolor{oursrow} \oursname{\systemshort{} (GPT)} & \textbf{98.6} & \textbf{84.1} & \textbf{94.0} & 87.7 & 34/22/1 & 35/21/1 \\
Codex (GPT) & 97.5 & 79.1 & 84.8 & 89.1 & 29/25/3 & 33/20/4 \\
\midrule
\multicolumn{7}{l}{\tabgroup{All (228 tasks)}} \\
\rowcolor{oursrow} \oursname{\systemshort{} (DS)} & \textbf{96.6} & \textbf{85.0} & \textbf{84.5} & 87.9 & 123/94/11 & 129/88/11 \\
OpenDesign (DS) & 95.2 & 69.6 & 75.2 & 88.3 & 121/98/9 & 121/98/9 \\
JarvisHub (DS) & 87.6 & 68.0 & 68.5 & 65.4 & 82/137/9 & 94/125/9 \\
\cmidrule(lr){1-7}
\rowcolor{oursrow} \oursname{\systemshort{} (GPT)} & 97.3 & \textbf{92.5} & \textbf{90.2} & 87.7 & 140/83/5 & 147/76/5 \\
Codex (GPT) & \textbf{98.1} & 78.1 & 75.3 & 89.8 & 125/96/7 & 138/80/10 \\
\bottomrule
\end{tabularx}
\end{papertable}

\paragraph{Supplementary task-level outcomes.}
\label{app:task-outcomes}
Table~\ref{tab:scale-results} also reports, for completeness, a supplementary task-level outcome: a task passes when every requested category is filled, every constraint the specification marks as critical holds, and the page minima of presentations and reports are met. The \emph{all} column applies this test to every deliverable, including requested intermediate outputs, and the \emph{final} column to final deliverables only. Full-corpus pass fractions use all 228 tasks as the denominator; condition-specific counts use the tasks in that condition. Unmeasurable outcomes have four sources: a report or slide deck delivered in a format without a readable page count, so its page minimum cannot be checked, which is the most common source; an original file the evaluator cannot read; a run the harness recorded as an infrastructure failure; and a critical constraint that could not be judged. Any measured failure of these conditions makes the task fail; an unmeasurable task has no measured failure but lacks evidence for a pass. Unmeasurable tasks remain in the denominator.

\paragraph{Domain coverage.}
The domain-level comparison is reported in Figure~\ref{fig:domains} in the main text. Each domain has 12 tasks and 25 to 80 applicable constraints, so these rates describe breadth of performance rather than establishing a significant difference in every domain.

\FloatBarrier
\subsection{Robustness of the paired comparisons}
\label{app:robustness}

\paragraph{Paired uncertainty and growth with scale.}
Accounting for related tasks preserves the same-model advantage. All paired constraint-satisfaction gaps at 6, 14, and 24 artifacts have 95\% intervals above zero (Table~\ref{tab:bootstrap}). For the project-family-clustered bootstrap, the 152 distinct projects are resampled with replacement 4,000 times, the three scale variants of a family are drawn together, and intervals are 95\% percentile intervals. The point-estimate constraint-satisfaction gaps widen from 6 to 24 artifacts against each baseline, but the interval for this growth excludes zero only against JarvisHub. This distinction separates an advantage at a given scale from evidence that the advantage grows with scale.

\begin{papertable}[h]
\centering\tabfont
\setlength{\tabcolsep}{2.0pt}
\caption{\textbf{Paired constraint-satisfaction gaps with family-clustered 95\% bootstrap intervals (points).} Growth is the gap at 24 artifacts minus the gap at 6.}
\label{tab:bootstrap}
\begin{tabular}{lrrrrr}
\toprule
{\tabhead Pair} & {\tabhead All tasks} & {\tabhead 6 art.} & {\tabhead 14 art.} & {\tabhead 24 art.} & {\tabhead Growth} \\
\midrule
\begin{tabular}[c]{@{}l@{}}\system\\ (DeepSeek) $-$ JarvisHub\end{tabular} & \papertableinterval{17.0}{12.4}{22.2} & \papertableinterval{10.0}{0.3}{19.6} & \papertableinterval{20.1}{7.2}{34.9} & \papertableinterval{31.3}{18.6}{45.8} & \papertableinterval{21.3}{3.8}{40.0} \\
\begin{tabular}[c]{@{}l@{}}\system\\ (DeepSeek) $-$ OpenDesign\end{tabular} & \papertableinterval{15.4}{9.4}{21.5} & \papertableinterval{15.7}{2.1}{30.8} & \papertableinterval{23.3}{9.9}{35.5} & \papertableinterval{25.1}{12.3}{38.4} & \papertableinterval{9.4}{$-5.3$}{24.0} \\
\begin{tabular}[c]{@{}l@{}}\system\\ (GPT) $-$ Codex\end{tabular} & \papertableinterval{14.4}{9.3}{19.3} & \papertableinterval{12.3}{6.0}{20.0} & \papertableinterval{17.9}{10.2}{26.4} & \papertableinterval{23.0}{9.0}{36.4} & \papertableinterval{10.6}{$-3.5$}{24.7} \\
\bottomrule
\end{tabular}
\end{papertable}

\paragraph{Sensitivity to missing deliveries.}
The same-model advantage is not explained by counting wholly missing output categories as unsatisfied constraints. Table~\ref{tab:scope-empty} repeats the comparison after excluding those constraints. A constraint none of whose target categories contains a delivered file counts as unsatisfied (Appendix~\ref{app:constraint-eligibility}); excluding such constraints, 2 to 39 per configuration, changes the same-model gaps over all tasks by at most 1.9 points, and every paired gap at 14 and 24 artifacts still excludes zero. The growth of the gap from 6 to 24 artifacts becomes 10.0 points [$-4.2$, 24.3] against Codex and 18.9 [0.8, 37.6] against JarvisHub, so the growth interval still excludes zero against JarvisHub but not against Codex. The growth against JarvisHub shrinks mainly because of its 12 such constraints at 24 artifacts, which narrow the gap there from 31.3 to 26.9 points.

\begin{papertable}[h]
\centering\tabfont
\setlength{\tabcolsep}{4pt}
\caption{\textbf{Constraint satisfaction (\%) without constraints on wholly undelivered categories.} Constraints none of whose target categories contains a delivered file are removed from numerator and denominator; parentheses give how many were removed. Reported rates, which count them as unsatisfied, are in Tables~\ref{tab:main-results} and~\ref{tab:scale-results}.}
\label{tab:scope-empty}
\begin{tabular}{lrrrr}
\toprule
{\tabhead System} & {\tabhead All tasks} & {\tabhead 6 art.} & {\tabhead 14 art.} & {\tabhead 24 art.} \\
\midrule
\rowcolor{oursrow} \oursname{\systemshort{} (DS)} & 86.2 (12) & 87.5 (2) & 83.7 (3) & 85.7 (0) \\
OpenDesign (DS) & 70.3 (9) & 70.6 (0) & 61.1 (6) & 61.0 (1) \\
JarvisHub (DS) & 71.1 (39) & 79.5 (6) & 64.0 (5) & 58.8 (12) \\
\midrule
\rowcolor{oursrow} \oursname{\systemshort{} (GPT)} & 92.7 (2) & 98.0 (1) & 94.2 (0) & 91.9 (0) \\
Codex (GPT) & 78.5 (5) & 85.1 (0) & 76.3 (0) & 68.9 (0) \\
\bottomrule
\end{tabular}
\end{papertable}

\paragraph{Sensitivity to judge abstention.}
Table~\ref{tab:abstention} tests whether judge abstentions change the reference-fidelity and identity-consistency ranking on the 228 tasks. Excluding them raises \system\ identity consistency to 90.1\% and 93.4\%. The baselines also improve, but remain below \system\ on both measures within each model group. The ranking is therefore the same under either treatment of unavailable pair evidence.

The table reports both denominator choices; most abstentions concern pairs whose views are not comparable, such as a plan drawing against a photograph, or a whole object against a detail.

\begin{papertable}[h]
\centering\tabfont
\setlength{\tabcolsep}{4pt}
\caption{\textbf{Judged image-pair measures with judge abstentions counted as mismatches (reported) and excluded.} Abstention counts are attributed to the system's own evidence; for reference fidelity, abstentions attributed to the external reference are excluded in both columns.}
\label{tab:abstention}
\begin{tabular}{lrrrrrr}
\toprule
& \multicolumn{3}{c}{\tabhead Identity consistency} & \multicolumn{3}{c}{\tabhead Reference fidelity} \\
\cmidrule(lr){2-4}\cmidrule(lr){5-7}
{\tabhead System} & {\tabhead Abstained} & {\tabhead Counted} & {\tabhead Excluded} & {\tabhead Abstained} & {\tabhead Counted} & {\tabhead Excluded} \\
\midrule
\rowcolor{oursrow} \oursname{\systemshort{} (DS)} & 41/656 & 84.5 & 90.1 & 137/1407 & 75.6 & 83.7 \\
OpenDesign (DS) & 93/621 & 75.2 & 88.4 & 351/1396 & 55.0 & 73.5 \\
JarvisHub (DS) & 38/600 & 68.5 & 73.1 & 199/1326 & 49.6 & 58.4 \\
\midrule
\rowcolor{oursrow} \oursname{\systemshort{} (GPT)} & 23/663 & 90.2 & 93.4 & 113/1414 & 79.3 & 86.2 \\
Codex (GPT) & 97/661 & 75.3 & 88.3 & 195/1425 & 71.3 & 82.6 \\
\bottomrule
\end{tabular}
\end{papertable}

\paragraph{Diagnostic: partial coverage of factual constraints.}
\label{sec:constraint-loss}
The diagnostic tests whether unmet constraints are absent from every target or fulfilled only in part of the delivered set. Of Codex's 191 unmet constraints with per-file records, 117 hold in at least half of their targets: many failures are omissions within a delivered set. Table~\ref{tab:constraint-targets} decomposes Equation~\ref{eq:dc-cs} to distinguish partial coverage from satisfaction in every target. Each factual constraint $c$ has a criterion $q_c(y)$ and target files $T(c)$, as defined in Appendix~\ref{app:measures}. \emph{Per-file constraint coverage} pools the successful constraint--target checks across the evaluated tasks:
\[
\frac{\sum_{(c,y)\in\mathcal{E}}\mathbf{1}[q_c(y)=1]}{|\mathcal{E}|},
\]
where $\mathcal{E}$ contains the evaluated constraint--target pairs with per-file records. Constraint-specific checks and target eligibility follow Appendix~\ref{app:constraint-eligibility}; text normalization and transcription exclusions apply as above. The denominator counts pairs, not distinct files: a file targeted by two constraints contributes two checks. A factual constraint satisfied in two of three target files gives $2/3$ coverage but an unsatisfied constraint; wholly undelivered categories are handled separately above.

On GPT-5.6 Sol, \system's per-file constraint coverage is 97.9--99.1\% across scales, versus 86.1--88.6\% for Codex (Figure~\ref{fig:constraint-coverage}). Mean targets per constraint increase from about 2.2 to 3.4 to 5.7, averaging the five system means; Table~\ref{tab:constraint-targets} gives each system's values. No system's coverage declines by more than 5 points from 6 to 24 artifacts, but each constraint must hold across more files. This diagnostic describes final-output coverage. Constraint satisfaction additionally depends on how failed checks are distributed among constraints; neither statistic identifies when a constraint was lost during execution.

\begin{papertable}[h]
\centering\tabfont
\setlength{\tabcolsep}{4pt}
\caption{\textbf{Constraint satisfaction decomposed by target file.} Columns give mean target files per constraint, per-file constraint coverage (\%), and constraint satisfaction (\%) at 6 / 14 / 24 artifacts.}
\label{tab:constraint-targets}
\begin{tabular}{lccc}
\toprule
{\tabhead System} & {\tabhead Targets per constraint} & {\tabhead Per-file coverage} & {\tabhead Constraint satisfaction} \\
\midrule
\rowcolor{oursrow} \oursname{\systemshort{} (DS)} & 2.21 / 3.38 / 5.76 & 87.9 / 89.2 / 97.0 & 86.4 / 82.1 / 85.7 \\
OpenDesign (DS) & 2.32 / 3.42 / 5.84 & 74.7 / 76.4 / 75.3 & 70.6 / 58.7 / 60.6 \\
JarvisHub (DS) & 2.14 / 3.29 / 5.30 & 90.2 / 86.3 / 85.5 & 76.3 / 61.9 / 54.4 \\
\midrule
\rowcolor{oursrow} \oursname{\systemshort{} (GPT)} & 2.21 / 3.47 / 5.68 & 99.1 / 98.3 / 97.9 & 97.4 / 94.2 / 91.9 \\
Codex (GPT) & 2.31 / 3.37 / 5.92 & 87.3 / 88.6 / 86.1 & 85.1 / 76.3 / 68.9 \\
\bottomrule
\end{tabular}
\end{papertable}

\begin{paperfigure}[t]
\centering
\includegraphics[width=\linewidth]{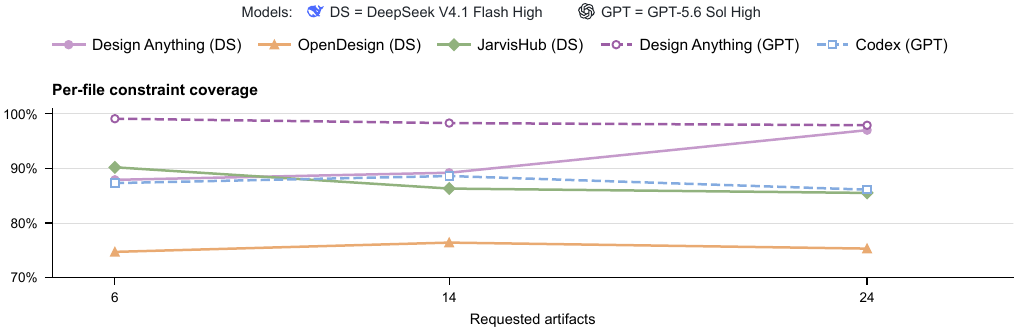}
\caption{\textbf{High per-file coverage can coexist with unmet constraints across a set.} Curves compare five configurations on 38 matched families, showing successful factual constraint--target checks as a percentage of evaluated pairs at each scale. Filled markers and solid lines denote DeepSeek V4.1 Flash High; hollow markers and dashed lines denote GPT-5.6 Sol High. A constraint passes only if every eligible target satisfies it.}
\label{fig:constraint-coverage}
\end{paperfigure}

\subsection{Reference coverage}
\label{app:reference-coverage}
\label{sec:reference-coverage}
We compare recorded reference reuse on the 228 DeepSeek tasks and examine its association with identity consistency. On these tasks, \system\ uses references in 95.3\% of image generations, versus 39.4\% for JarvisHub. Reference coverage measures this share of generation requests. The recorded LUMA project illustrates how a selected concept feeds later boards, video, and a presentation (Figure~\ref{fig:luma-case}); here we quantify reference use across runs.

For \system, references are read from the generation requests recorded in each run; for JarvisHub, from the reference edges of its exported canvas graph. Across these tasks, \system\ issues 2{,}680 image generations and 4 tasks fall below one-half reference coverage; JarvisHub issues 4{,}258 and 130 tasks fall below one half (three JarvisHub tasks produced no image). On the 24-artifact tasks, coverage reaches 97.9\% for \system\ and 40.2\% for JarvisHub; 74.9\% of the referenced generations of \system\ take an image generated earlier in the same run.

Figure~\ref{fig:refcov}b examines reference reuse and identity consistency within JarvisHub's 103 tasks with identity pairs; \system's coverage is close to its ceiling in every condition. Higher coverage is associated with higher identity consistency (Spearman $\rho=0.40$, $p<0.001$), including after ranking within conditions ($\rho=0.47$, $p<0.001$). The associations with set consistency, constraint satisfaction, and reference fidelity are weaker. Reference reuse is thus most closely associated with object identity in this diagnostic; observed coverage does not isolate the workflow graph's causal contribution.

Identity consistency per task is the pair-level rate within a run, the same quantity that Table~\ref{tab:main-results} pools. Tasks have a median of five identity pairs, so per-task values are coarse and 46 of the 103 tasks sit at 0 or 100\%. The identity correlation is positive in every condition: 0.73 at 6 artifacts ($n=17$), 0.53 at 14 ($n=18$), 0.60 at 24 ($n=16$), 0.37 on breadth tasks ($n=32$), and 0.13 on reference-material tasks ($n=20$). Ranks taken within each condition are pooled for the adjusted correlation above; correlations with set consistency, constraint satisfaction, and reference fidelity over all tasks are 0.26, 0.10, and 0.05.

\begin{paperfigure}[h]
\centering
\includegraphics[width=\linewidth]{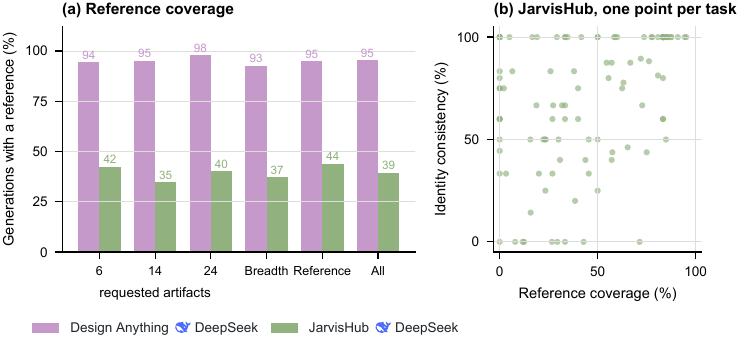}
\caption{\textbf{\system\ uses references more often; reference use is associated with identity consistency within JarvisHub.} Panel (a) shows the share of image generations with at least one reference input, pooled over each condition's tasks, for \system\ and JarvisHub on DeepSeek V4.1 Flash High. Breadth and Reference denote breadth and reference-material tasks. Panel (b) plots JarvisHub's reference coverage against identity consistency, with one point per task with identity pairs (Spearman $\rho=0.40$, $n=103$).}
\label{fig:refcov}
\end{paperfigure}

\subsection{Resource use in the main benchmark}
\label{sec:efficiency}
LLM input volume increases with project scale under both configurations. Table~\ref{tab:tokens} reports median input tokens per task for \system: on DeepSeek, the median rises from 1.4M at 6 artifacts to 6.6M at 14 and 9.1M at 24; on GPT, the orchestrating agent alone reads 5.6M, 16.6M, and 18.4M. The controlled delivery-loop cost comparison is in Appendix~\ref{app:continuation-ablation}.

The DeepSeek counts come from per-task token records; the GPT counts come from each run's records and cover the orchestrating agent only, excluding helper sessions, so the two columns are not directly comparable. Because token records also differ in scope across harnesses, we do not compare tokens or cost across systems.

\begin{papertable}[h]
\centering\tabfont
\caption{\textbf{Median LLM input tokens per task for \system\ (millions).} The GPT-5.6 Sol High column counts the orchestrating agent only.}
\label{tab:tokens}
\begin{tabular}{lrr}
\toprule
{\tabhead Condition} & {\tabhead DeepSeek V4.1 Flash High} & {\tabhead GPT-5.6 Sol High} \\
\midrule
6 artifacts & 1.36 & 5.59 \\
14 artifacts & 6.58 & 16.57 \\
24 artifacts & 9.06 & 18.38 \\
Breadth & 4.22 & 16.42 \\
Reference materials & 2.37 & 12.34 \\
\midrule
All & 4.04 & 14.63 \\
\bottomrule
\end{tabular}
\end{papertable}

\subsection{Design quality}
\label{app:project-design-quality}
Table~\ref{tab:pdq} compares five configurations on four delivery-weighted design-quality dimensions across the 228 tasks. \system{} leads on all four under DeepSeek and scores 88.5 on aesthetics and 92.5 on domain criteria under GPT. Codex leads on GPT technical quality and text rendering by 0.1 and 0.5 points, respectively, although differences this small are below the observed judging variability.

\begin{papertable}[h]
\centering\tabfont
\caption{\textbf{Design quality (100-point scale).} Scores combine four image-checklist dimensions with project delivery coverage (Equation~\ref{eq:project-design-quality}).}
\label{tab:pdq}
\begin{tabular}{lrrrr}
\toprule
{\tabhead System} & {\tabhead Technical} & {\tabhead Aesthetics} & {\tabhead Text} & {\tabhead Domain} \\
\midrule
\systemshort{} (DS) & 88.4 & 87.1 & 93.0 & 92.0 \\
\systemshort{} (GPT) & 88.8 & 88.5 & 93.8 & 92.5 \\
OpenDesign (DS) & 85.1 & 82.3 & 80.6 & 85.5 \\
JarvisHub (DS) & 80.9 & 80.2 & 84.7 & 84.2 \\
Codex (GPT) & 88.9 & 87.9 & 94.3 & 90.8 \\
\bottomrule
\end{tabular}
\end{papertable}

\paragraph{Scoring coverage.}
All 228 tasks of every configuration were judged under the protocol in Appendix~\ref{app:measures}. The judged images per configuration are 2{,}089 for \systemshort{} (DS), 2{,}102 for \systemshort{} (GPT), 2{,}024 for OpenDesign, 1{,}920 for JarvisHub, and 2{,}126 for Codex; at most four images per configuration were left unscored after a failed judge call.

\paragraph{Judge sensitivity.}
Claude Opus 4.8 rescored the same delivered images on a stratified 15-task sample (three per condition) for \systemshort{} (GPT), OpenDesign, and JarvisHub; Codex was not included. This check tests the image-checklist component of design quality. On the sample's checklist scores, before weighting by delivery completion, both judges place \system{} and JarvisHub within 2.4 points of each other on every dimension and above OpenDesign, with per-task Spearman correlations of 0.37--0.60 (the DeepSeek scores for \systemshort{} (GPT) come from an earlier pass of the same judge). This analysis characterizes judge sensitivity of the checklist component rather than whole-set designer preference.

\clearpage
\section{Ablation Studies}
\label{app:ablations}
The studies below examine the delivery loop, planning interfaces, and state summaries. In the delivery-loop study, Stop at model end versus Generic prompt tests prompting after a normal stop within the same reduced harness; Generic prompt versus Complete compares a bundle of mechanisms and its execution efficiency. The planning and state-summary studies examine narrower changes separately. The delivery-loop study uses 19 tasks from the main corpus; planning and state-summary studies use separate tasks and harness versions. Each study reports its own configuration and contributes no observations to the 228-task benchmark aggregates. No study uses skills written for a particular design domain.

\subsection{Delivery loop and execution efficiency}
\label{app:continuation-ablation}
This study asks whether replying after the model stops restores delivery, and how the complete harness changes repeated generation and cost. It uses the 24-artifact version of one base family per domain (19 tasks), the family whose brief asks the system to propose alternative directions and select one, with the LLM and image-generation configuration of the main comparison. Each configuration has one run per task, including a fresh Complete run separate from the main comparison. The verifier is enabled for Complete and disabled for the two reduced configurations. All three are scored with the same evaluator. Approval requests raised through the checkpoint tool receive the same scripted answers in every configuration; a question written into a final message ends the turn, and Stop at model end provides no reply.

\begin{papertable}[h]
\centering\tabtextfont
\caption{\textbf{Delivery-loop configurations.} The first two configurations differ only in a generic reply after stopping within the reduced harness. Complete restores the full set of mechanisms.}
\label{tab:continuation-config}
\begin{tabularx}{\linewidth}{@{}lYYY@{}}
\toprule
{\tabhead Mechanism} & {\tabhead Stop at model end} & {\tabhead Generic prompt} & {\tabhead Complete} \\
\midrule
After the model stops & End run & Fixed prompt to keep working & Delivery loop (Section~\ref{sec:method-control}) \\
Contract and planning tools & Removed & Removed & Retained \\
Task and progress reminders & Removed & Removed & Retained \\
Automatic verifier & Disabled & Disabled & Enabled \\
Batched asynchronous execution & Removed & Removed & Retained \\
\bottomrule
\end{tabularx}
\end{papertable}

\paragraph{Configuration definitions.}
\emph{Stop at model end} removes the delivery-loop controller, the contract and planning tools, batched asynchronous execution, and every channel that restates the task or its progress, such as reminders of the original brief and progress lines in the status message. Synchronous execution, read tools, context compaction, scripted approval answers, and retries after transport failures remain, since a generic tool-using harness has them. A run ends when the model ends a turn without calling a tool. \emph{Generic prompt} keeps every removal and, whenever the model ends a turn, sends the fixed user message ``Continue working until everything the original brief asks for is fully delivered. If everything is already delivered, say so briefly and make no canvas changes.'' The message names no outstanding item. The loop stops after two consecutive rounds without canvas change, after 100 turns, or at the 150-minute wall-clock limit. \emph{Complete} is the full harness of Section~\ref{sec:method}. Stop at model end versus Generic prompt tests the addition of the fixed prompt; Generic prompt versus Complete compares the combined mechanisms in Table~\ref{tab:continuation-config}.

\paragraph{Delivery and constraint outcomes.}
Table~\ref{tab:continuation} compares three configurations on the 19 paired 24-artifact tasks. Complete achieves 96.1\% delivery and 90.4\% constraint satisfaction. It improves both over Stop at model end ($p=0.018$ and $p=0.001$), but its delivery and constraint differences from Generic prompt are not statistically significant ($p=0.39$ and $p=0.063$). This supports fulfillment gains over stopping unaided on larger projects; the comparison with Generic prompt does not isolate one component.

Paired tests are two-sided exact Wilcoxon signed-rank tests paired by task; tasks with equal values are excluded and tied absolute differences receive average ranks. Against the complete harness, Stop at model end delivers less on 8 tasks and more on 3 and satisfies fewer constraints on 11 tasks and more on none. Generic prompt delivers more on 7 tasks and less on 4, with 8 ties; constraint satisfaction favors Complete on 6 tasks and Generic prompt on 1.

\paragraph{Early stops and constraint outcomes.}
The ``Runs $<50\%$'' column counts runs delivering fewer than half of their requested artifacts. Six Stop at model end runs fall below half delivery against no Complete run (Fisher exact test, two-sided $p=0.020$, one-sided $p=0.010$). All six end within three turns: three ask for confirmation of a direction or a costly next stage, two announce further deliverables, and one exhausts the output-token limit before building anything. Generic prompt recovers each to 23 or 24 deliverables.

Missing deliveries explain only part of the constraint gap. Constraints on wholly undelivered categories count as unsatisfied (Appendix~\ref{app:constraint-eligibility}); these six runs account for at most 12 of the 24 fewer constraints satisfied relative to Complete. At least 12 fewer satisfied constraints therefore occur in runs that deliver 20 or more of their 24 artifacts.

\paragraph{Repeated generation.}
On the same 19 tasks (Table~\ref{tab:continuation}), the generic configuration issues more generation requests than the complete harness on 15 tasks and fewer on 3 ($p=0.002$); counting only completed requests gives 14 and 3 ($p=0.001$). The extra requests come after the deliverables are largely in place, when the prompt keeps the agent regenerating delivered work; in one task the generic configuration issues 68 requests against 27. This supports an efficiency advantage of the full harness on these larger projects, alongside the cost comparison below.

\begin{papertable}[h]
\centering\tabfont
\caption{\textbf{Delivery-loop study: mean cost per run on 17 paired tasks.} Two Generic prompt runs reached the wall-clock limit without step or token counts and are excluded with their pairs. Input tokens (millions) and LLM cost include sub-agents; cost uses list prices.}
\label{tab:continuation-cost}
\begin{tabular}{lrrr}
\toprule
{\tabhead Measurement} & {\tabhead Stop at model end} & {\tabhead Generic prompt} & {\tabhead Complete} \\
\midrule
Agent steps & 31.2 & 79.7 & 36.4 \\
Input tokens (M) & 16.9 & 41.8 & 17.9 \\
LLM cost (USD) & 0.48 & 1.30 & 0.43 \\
Generation requests & 18.3 & 37.1 & 23.8 \\
Delivery completion (\%) & 72.1 & 96.8 & 96.1 \\
\bottomrule
\end{tabular}
\end{papertable}

\paragraph{LLM cost.}
On the 17 paired tasks with complete telemetry, Complete costs \$0.43 per run versus \$1.30 for Generic prompt (Table~\ref{tab:continuation-cost}). The generic configuration takes more agent steps on all 17 tasks, and more input tokens and LLM cost on 15 of 17 ($p<0.001$ for each). On the same 17 tasks we detect no difference between the Stop at model end configuration and the complete harness in steps (8 higher, 9 lower), input tokens (7 and 10), or cost (9 and 8). Mean LLM costs are \$0.48 and \$0.43, respectively; this sample does not establish cost equivalence.

\subsection{Planning-interface study}
\label{app:planning-study}
This study asks whether structured planning interfaces improve delivery when other channels for tracking requirements remain available. It compares three planning interfaces. \emph{Structured} retains the structured project-plan and workflow-construction tools. \emph{Text plan} replaces those tools with an instruction to maintain a free-text deliverable checklist. \emph{Planning tools off} removes them without a replacement instruction. Delivery groups, outstanding-work messages, direct canvas reads, status tools, retained history, and new model turns remain available in every configuration. The automatic verifier is disabled in all three configurations, including Structured, to keep its availability fixed.

On eight tasks, delivery completion is 7.7 points lower with Planning tools off than with Structured, and 2.1 points lower with Text plan. These comparisons measure the incremental contribution of structured planning interfaces alongside the other channels for tracking the task. Those retained channels offer a possible explanation for the modest gaps.

\subsection{State-summary efficiency}
\label{app:digest-study}
The state-summary study tests whether the summary reduces direct inspection of the canvas. It removes the summary while retaining direct read tools and measures the fraction of tool calls devoted to active canvas reading. Nine of ten paired tasks read more without the summary; the geometric mean ratio of reading fractions is 2.06 (one-sided exact sign test, $p=0.011$).

The paired result shows a larger share of tool calls devoted to direct state inspection without the summary; the experiment does not measure final design quality.

\clearpage
\section{Qualitative Examples}
\label{app:cases}
The visual evidence complements the domain and scale results in Section~\ref{sec:overall}: Figures~\ref{fig:compare-app-1}--\ref{fig:compare-app-2} enlarge the matched comparison in Figure~\ref{fig:comparison}; Figure~\ref{fig:luma-case} traces reuse of a selected design within one 24-artifact project; and Table~\ref{tab:gallery-index} indexes the gallery by domain and requested artifact count. These recorded examples illustrate delivered work and its dependencies.

\subsection{Enlarged outputs for Figure~\ref{fig:comparison}}
Figures~\ref{fig:compare-app-1} and~\ref{fig:compare-app-2} enlarge the 14-artifact animation brief of Figure~\ref{fig:comparison} for all five configurations. Each block keeps the deliverable order, credit counts, and markers of Figure~\ref{fig:comparison}. Documents show one page and videos one frame; credit counts follow the evaluator. OpenDesign places two keyframes on each page and its three act storyboards on one page, so these images repeat across positions.

Keyframes S01--S06 are shots numbered in the brief. Red markers identify five content errors checked against the brief. In the S06 keyframe of \system\ (DeepSeek), each traveler still holds the other's case. JarvisHub's color script replaces the otter characters with humans; its complete storyboard departs from the fixed actions of S03, S04, and S06 and omits both dialogue lines; and its Act~3 storyboard replaces the wheeled cases with cardboard boxes. Codex's Act~2 storyboard repeats one image for both S03 phases. The markers are examples, not an exhaustive audit. A slashed circle means the evaluator did not credit the requested artifact, not that no file was exported; OpenDesign, for example, exported a shot table, a web-page animatic, and a presentation that the evaluator did not credit to these requested artifacts.

\begin{paperfigure}[p]
  \centering
  \includegraphics[width=\linewidth]{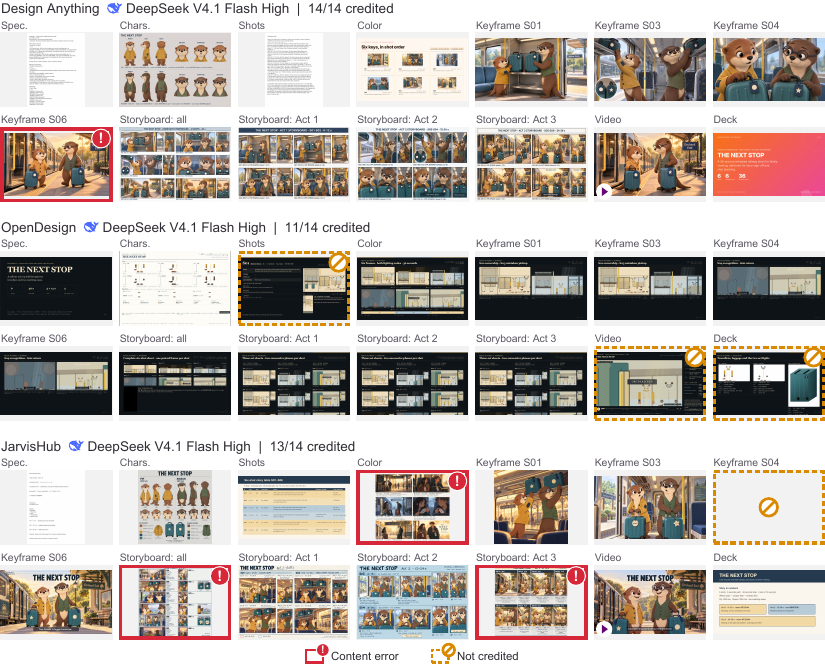}
  \caption{\textbf{Enlarged outputs for Figure~\ref{fig:comparison}: DeepSeek V4.1 Flash High.} Each block shows one configuration's 14 requested artifacts in deliverable order, with the markers of Figure~\ref{fig:comparison}.}
  \label{fig:compare-app-1}
\end{paperfigure}

\begin{paperfigure}[p]
  \centering
  \includegraphics[width=\linewidth]{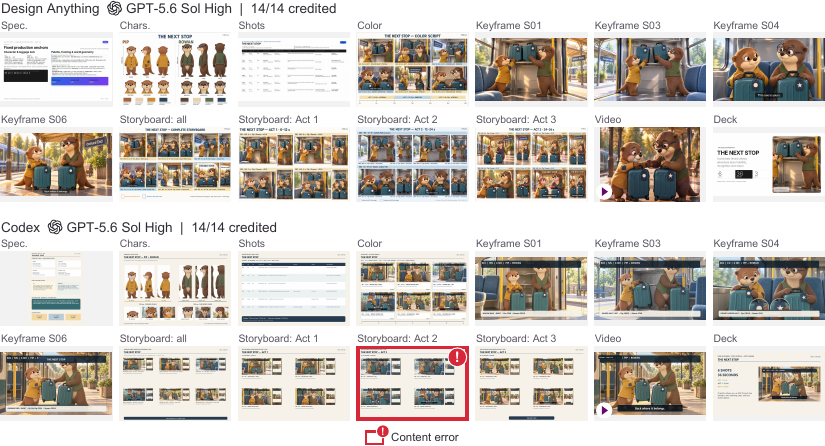}
  \caption{\textbf{Enlarged outputs for Figure~\ref{fig:comparison}: GPT-5.6 Sol High.} The layout follows Figure~\ref{fig:compare-app-1}.}
  \label{fig:compare-app-2}
\end{paperfigure}

\clearpage
\subsection{A recorded project and its shared dependencies}
This case asks how an earlier design decision becomes an input to later image, video, and presentation outputs within one 24-artifact project.

\begin{paperfigure}[!htb]
  \centering
  \includegraphics[width=\linewidth]{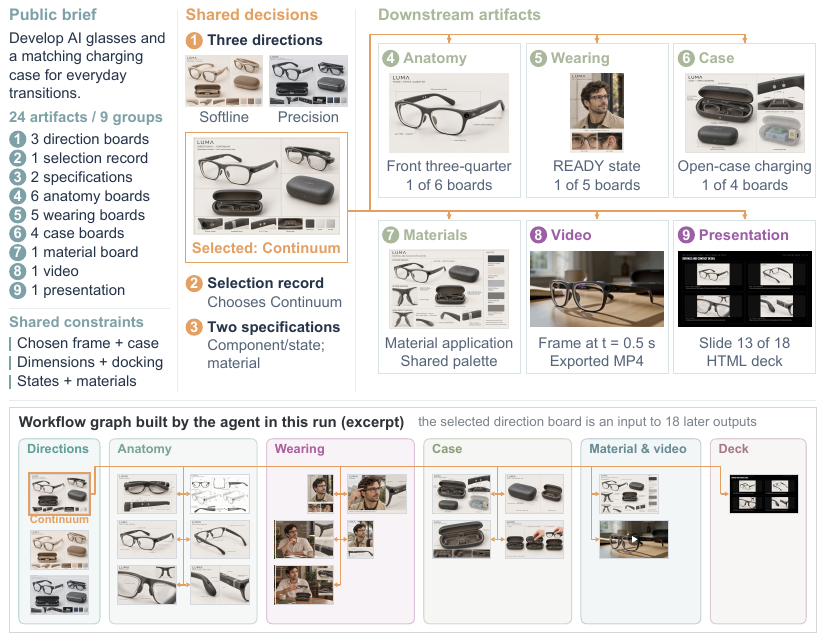}
  \caption{\textbf{Recorded cross-modal dependencies in the LUMA project.} This DeepSeek V4.1 Flash High task requests 24 artifacts in nine numbered groups. Selected real outputs show how one run reuses the chosen Continuum direction board across design boards, video, and a presentation; orange arrows mark recorded input connections. The bottom panel redraws an excerpt of the workflow graph that the agent built in this run: the 18 recorded connections that take the selected Continuum board as an input; text prompts, provided references, and the other 61 connections are omitted.}
  \label{fig:luma-case}
\end{paperfigure}

Figure~\ref{fig:luma-case} uses the LUMA product-design task from the main-comparison run of \system{} on DeepSeek V4.1 Flash High. Its public brief requests AI glasses and a matching charging case, with 24 artifacts in nine groups: three direction boards, one selection record, two specifications, six anatomy/detail boards, five wearing/use boards, four case boards, one material board, one video, and one presentation. Shared constraints concern the chosen product identity, dimensions and docking, operating states, and material choices.

The written selection record chooses Continuum from three directions. Its direction-board node has 18 recorded outgoing connections, which make it an input to all 16 anatomy, wearing, case, and material boards, to video generation, and to the presentation. The agent-built workflow graph has 79 connections in total. In the top panel, the orange arrows draw these 18 connections once per group; the bottom panel draws each of them.

This run delivers all 24 artifacts and satisfies 6 of its 7 constraints; the identity judge rates 16 of 18 image pairs as the same product. The unmet constraint shows that reuse alone does not guarantee full fulfillment of the brief.

\subsection{Agent-built workflow graphs and image boards of thirteen projects}
Figures~\ref{fig:showcase-1} to~\ref{fig:showcase-11} show thirteen more projects that we selected for display, each from one recorded run of \system{} with GPT-5.6 Sol High as the LLM. The gallery illustrates the range of domains and project scales handled by the same harness.

\begin{papertable}[h]
\centering\tabtextfont
\caption{\textbf{Gallery index.} Domains and requested artifact counts for thirteen recorded projects, selected for display.}
\label{tab:gallery-index}
\begin{tabularx}{\linewidth}{@{}lYl@{}}
\toprule
{\tabhead Figure} & {\tabhead Domain} & {\tabhead Artifacts} \\
\midrule
\ref{fig:showcase-1} & Fashion (first project) & 24 \\
\ref{fig:showcase-2} & Game art & 24 \\
\ref{fig:showcase-3} & Brand identity & 24 \\
\ref{fig:showcase-4} & Jewelry & 24 \\
\ref{fig:showcase-5} & Landscape design & 24 \\
\ref{fig:showcase-6} & Photography & 24 \\
\ref{fig:showcase-7} & Fashion (second project) & 24 \\
\ref{fig:showcase-8} & Packaging & 24 \\
\ref{fig:showcase-9} & UI/UX & 24 \\
\ref{fig:showcase-10} & Exhibition design; landscape design & 14 each \\
\ref{fig:showcase-11} & Furniture; product design & 6 each \\
\bottomrule
\end{tabularx}
\end{papertable}

\begin{paperfigure}[p]
  \centering
  \includegraphics[width=\linewidth]{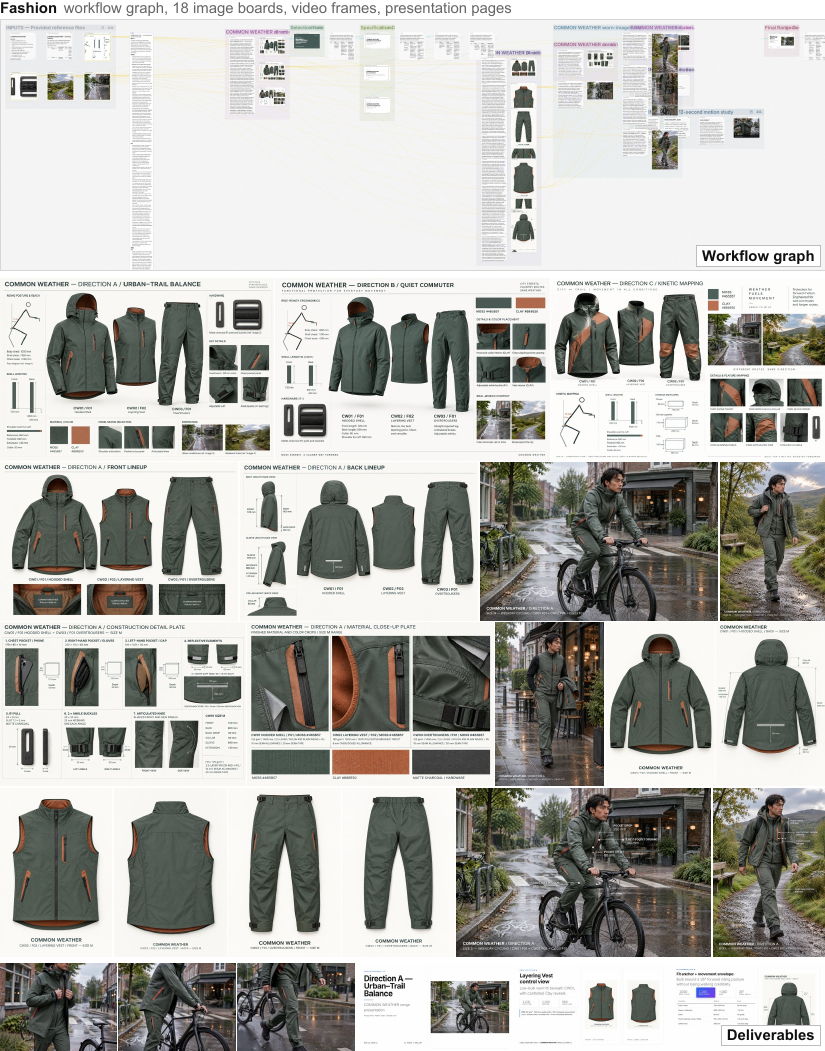}
  \caption{\textbf{A 24-artifact fashion project (I).} Each block shows, from top to bottom, the workflow graph the agent built during the run, its delivered image boards in deliverable order, and up to three video frames and three presentation pages. Projects were selected for display.}
  \label{fig:showcase-1}
\end{paperfigure}

\begin{paperfigure}[p]
  \centering
  \includegraphics[width=\linewidth]{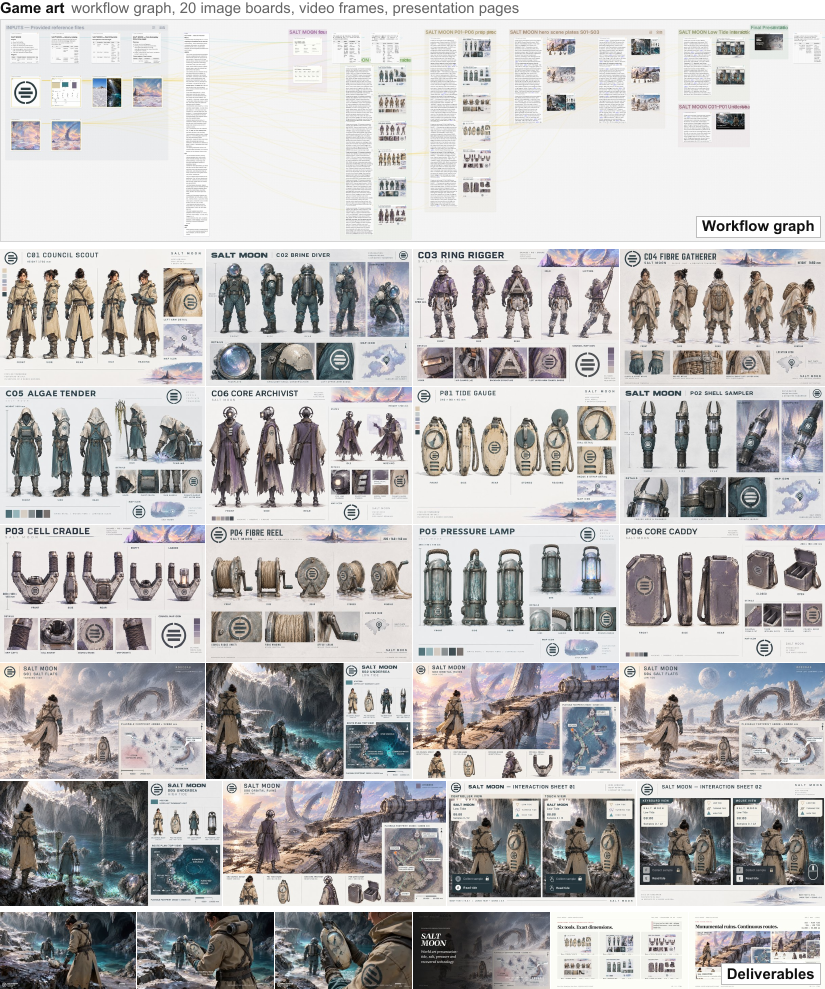}
  \caption{\textbf{A 24-artifact game-art project (II).} The layout follows Figure~\ref{fig:showcase-1}.}
  \label{fig:showcase-2}
\end{paperfigure}

\begin{paperfigure}[p]
  \centering
  \includegraphics[width=\linewidth]{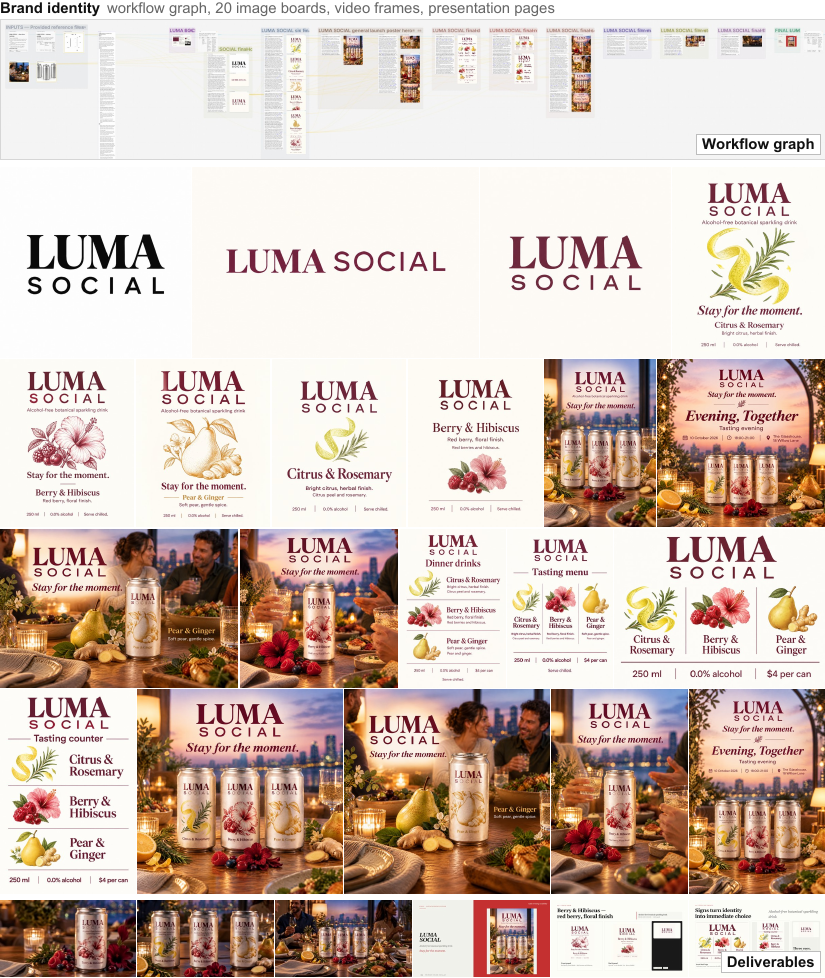}
  \caption{\textbf{A 24-artifact brand-identity project (III).} The layout follows Figure~\ref{fig:showcase-1}.}
  \label{fig:showcase-3}
\end{paperfigure}

\begin{paperfigure}[p]
  \centering
  \includegraphics[width=\linewidth]{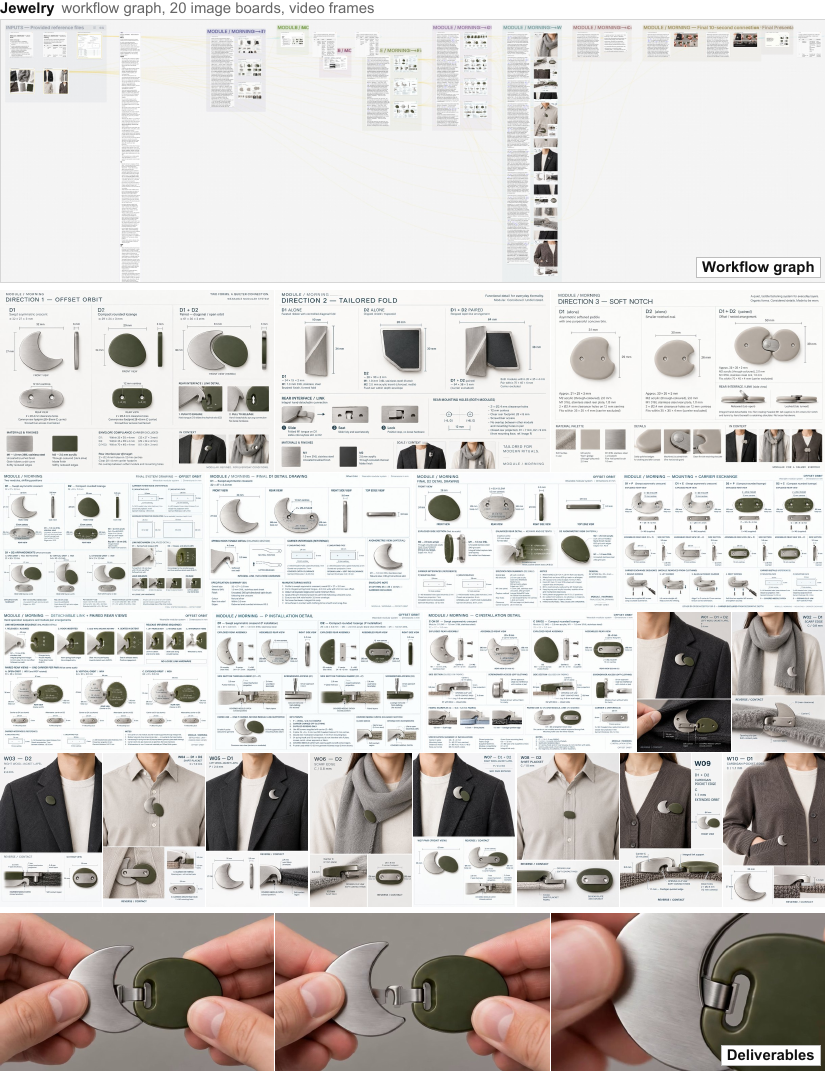}
  \caption{\textbf{A 24-artifact jewelry project (IV).} The layout follows Figure~\ref{fig:showcase-1}.}
  \label{fig:showcase-4}
\end{paperfigure}

\begin{paperfigure}[p]
  \centering
  \includegraphics[width=\linewidth]{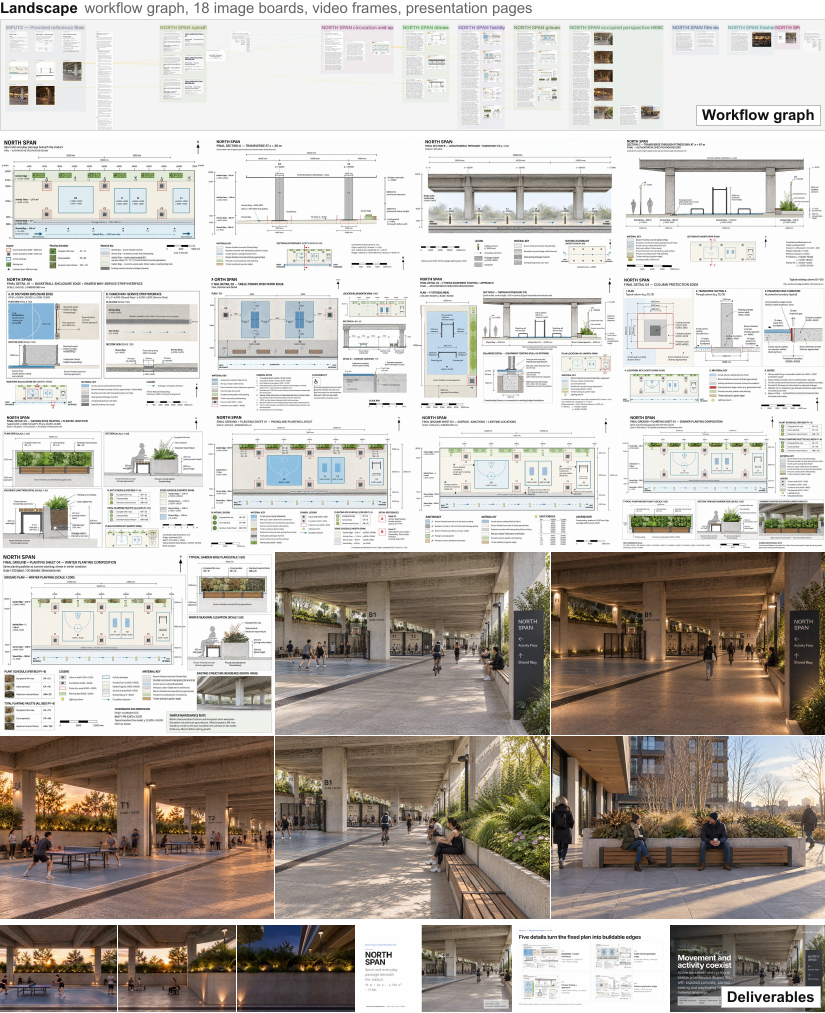}
  \caption{\textbf{A 24-artifact landscape-design project (V).} The layout follows Figure~\ref{fig:showcase-1}.}
  \label{fig:showcase-5}
\end{paperfigure}

\begin{paperfigure}[p]
  \centering
  \includegraphics[width=\linewidth]{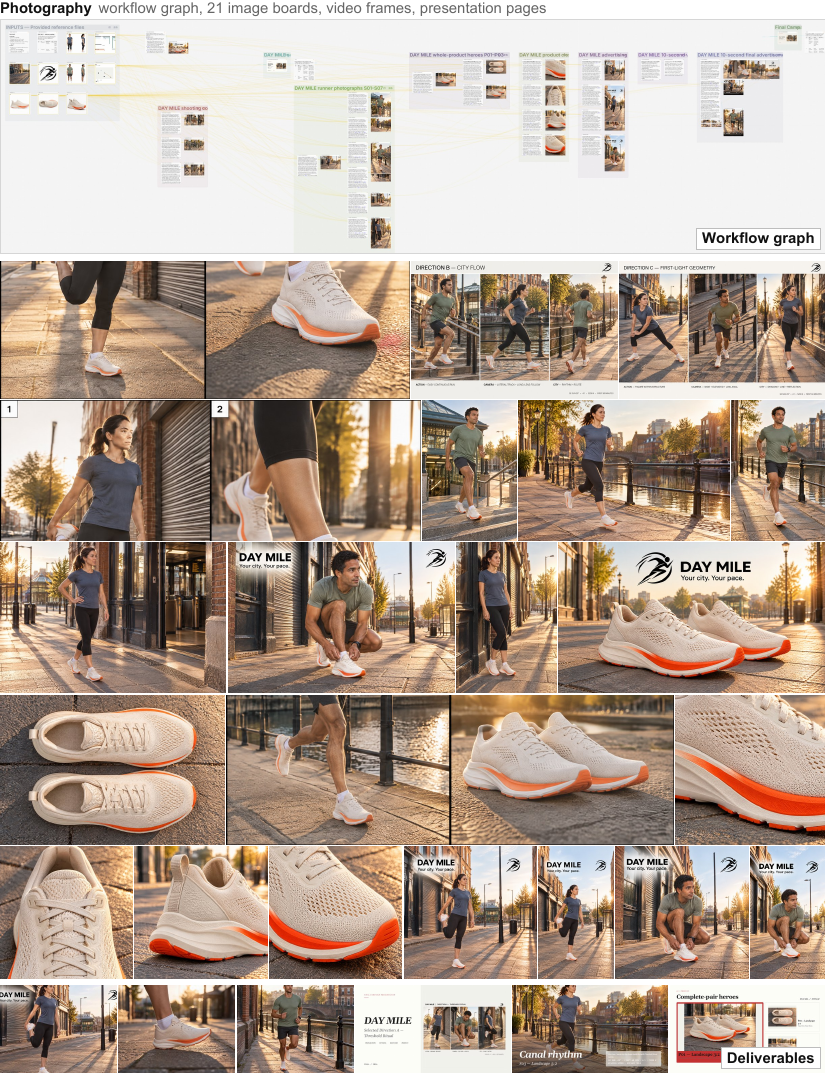}
  \caption{\textbf{A 24-artifact photography project (VI).} The layout follows Figure~\ref{fig:showcase-1}.}
  \label{fig:showcase-6}
\end{paperfigure}

\begin{paperfigure}[p]
  \centering
  \includegraphics[width=\linewidth]{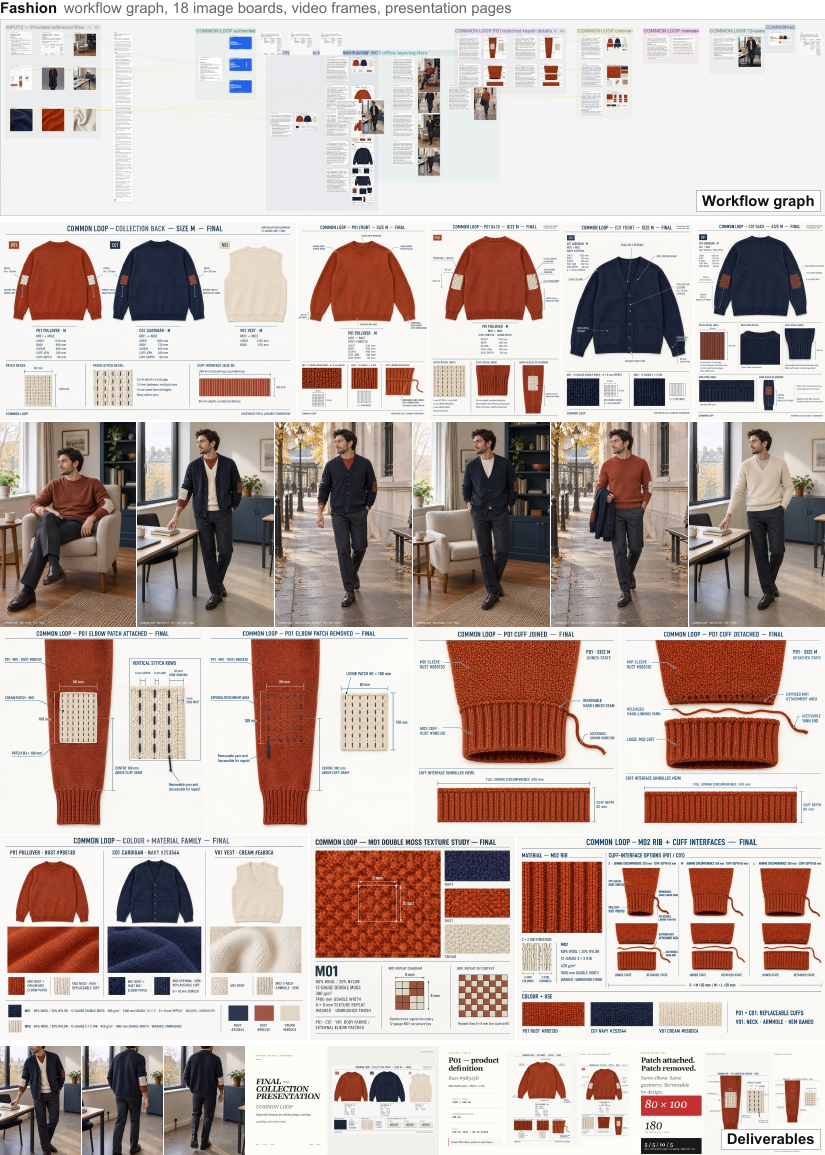}
  \caption{\textbf{A second 24-artifact fashion project (VII).} The layout follows Figure~\ref{fig:showcase-1}.}
  \label{fig:showcase-7}
\end{paperfigure}

\begin{paperfigure}[p]
  \centering
  \includegraphics[width=\linewidth]{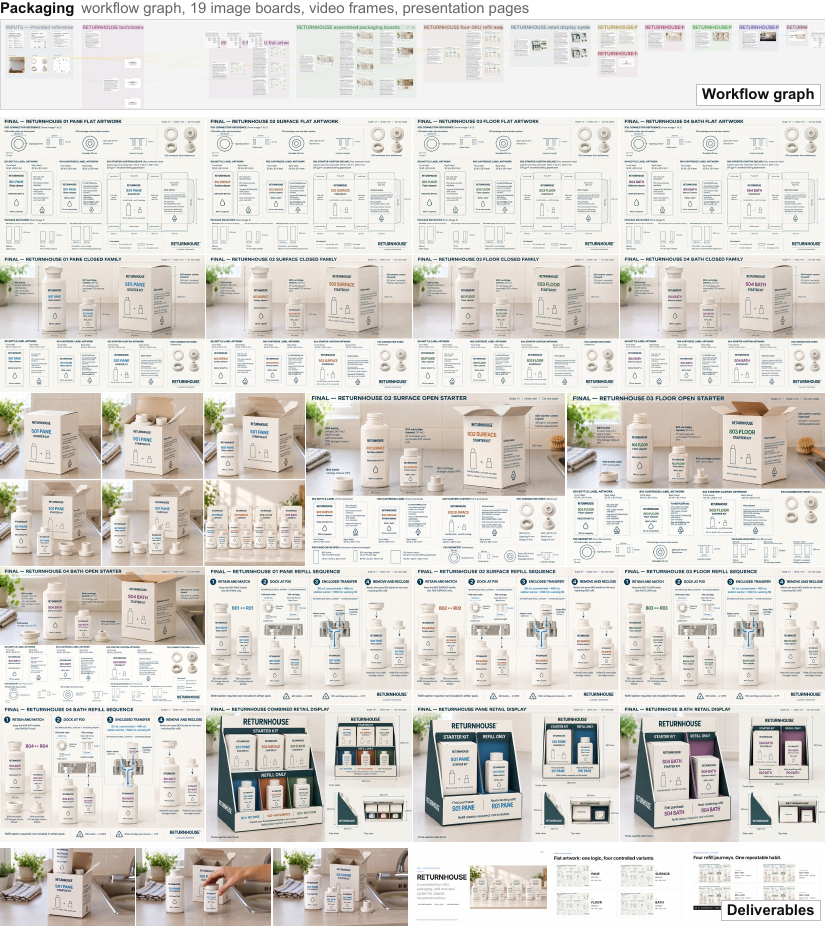}
  \caption{\textbf{A 24-artifact packaging project (VIII).} The layout follows Figure~\ref{fig:showcase-1}.}
  \label{fig:showcase-8}
\end{paperfigure}

\begin{paperfigure}[p]
  \centering
  \includegraphics[width=\linewidth]{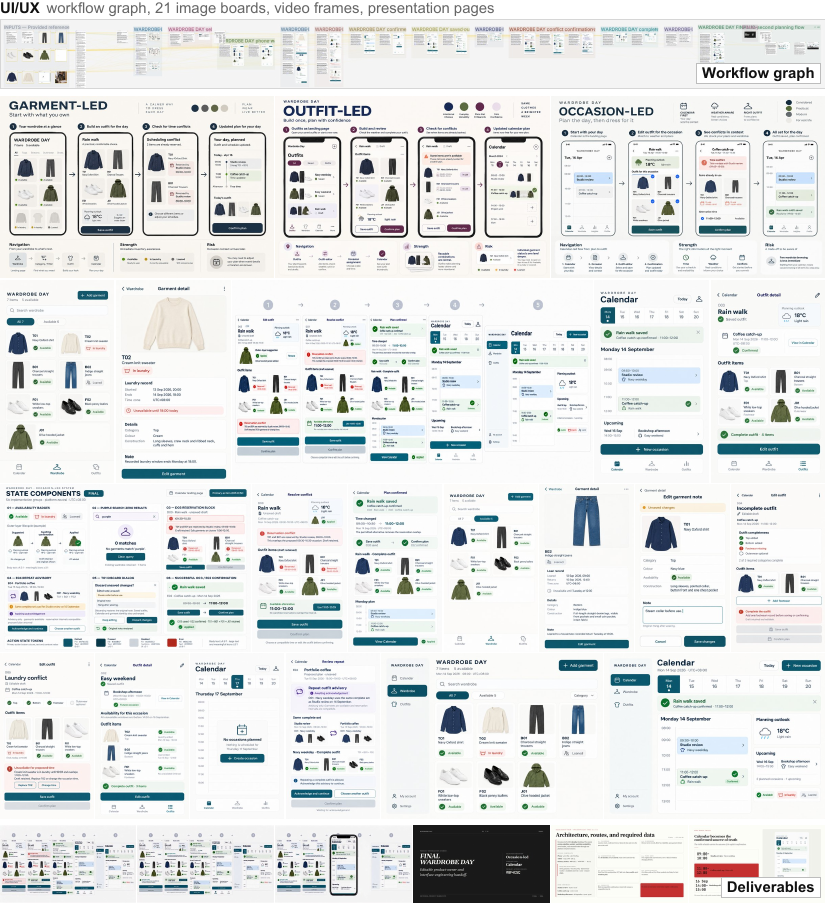}
  \caption{\textbf{A 24-artifact UI/UX project (IX).} The layout follows Figure~\ref{fig:showcase-1} and includes three direction boards.}
  \label{fig:showcase-9}
\end{paperfigure}

\begin{paperfigure}[p]
  \centering
  \includegraphics[width=\linewidth]{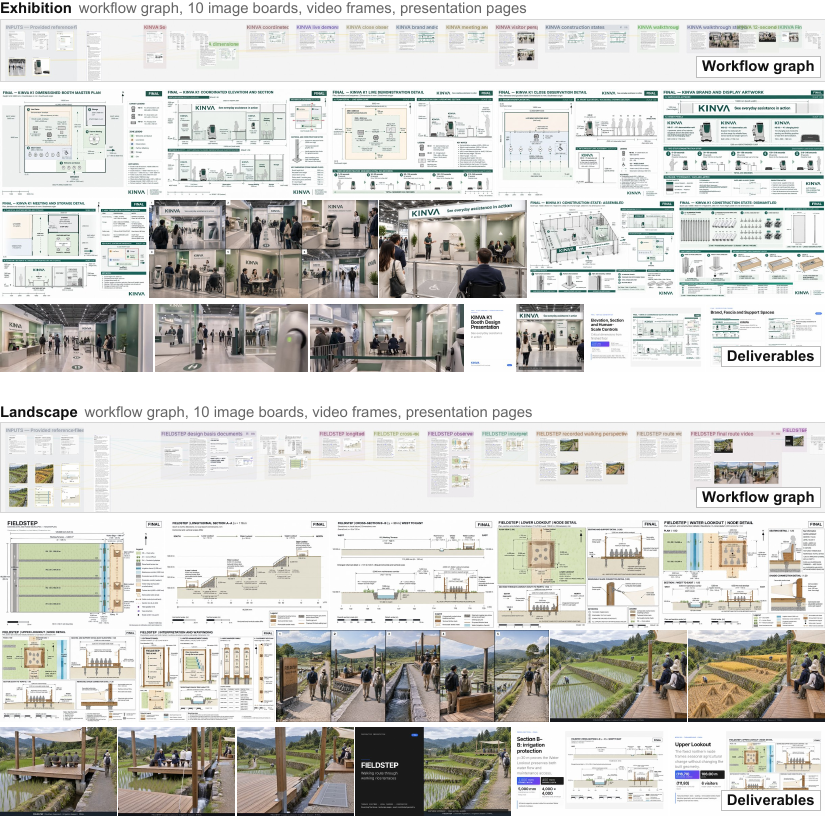}
  \caption{\textbf{Two 14-artifact projects (X).} The upper and lower blocks show exhibition design and landscape design, respectively. Both follow the layout of Figure~\ref{fig:showcase-1}.}
  \label{fig:showcase-10}
\end{paperfigure}

\begin{paperfigure}[p]
  \centering
  \includegraphics[width=\linewidth]{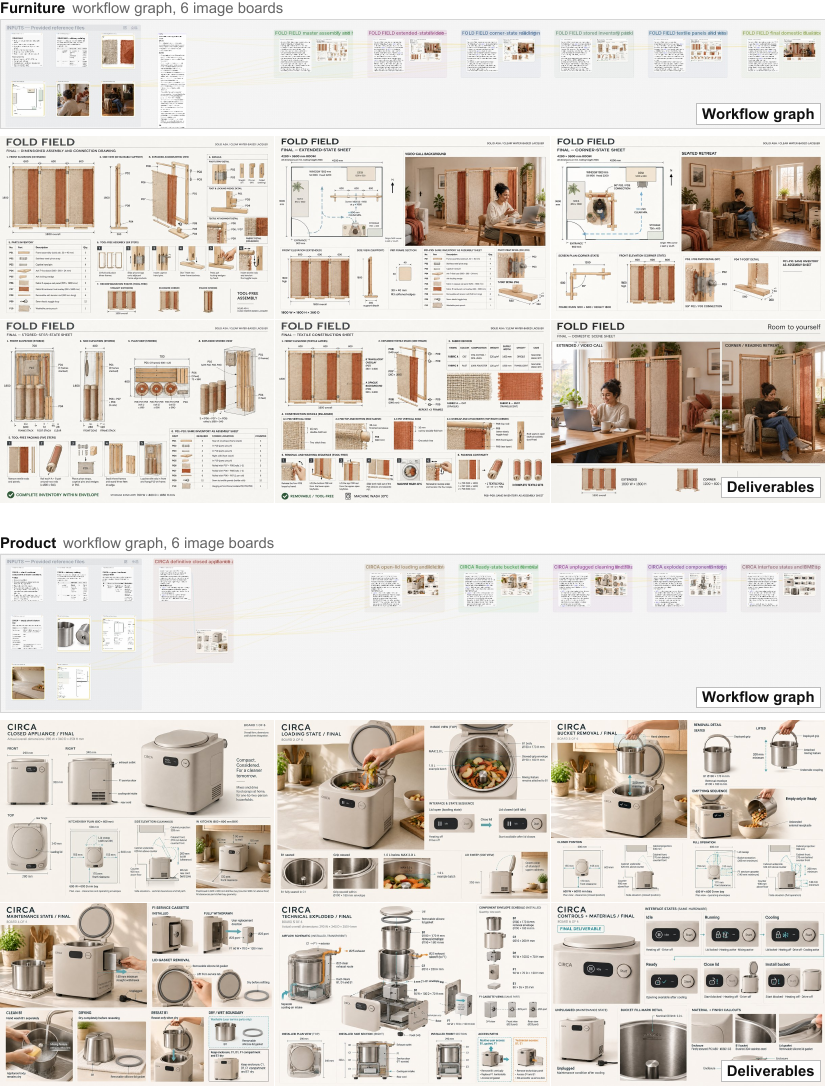}
  \caption{\textbf{Two 6-artifact projects (XI).} The upper and lower blocks show furniture and product design, respectively. Both follow the layout of Figure~\ref{fig:showcase-1}.}
  \label{fig:showcase-11}
\end{paperfigure}

\clearpage
\section{Limitations}
\label{app:limitations}
\paragraph{Evaluation and run variability.}
Each configuration contributes one final result per task; repeated runs would further characterize run-to-run variation. The image-checklist component of design quality relies on a vision-language judge. Whole-set designer preferences complement these measurements without identifying which properties drive preference.

\paragraph{Scope of system comparisons.}
The main comparisons hold the LLM fixed and evaluate complete systems, including their tools and execution policies. The delivery-loop comparison bundles several mechanisms; planning-interface and state-summary studies examine narrower changes within their own configurations (Appendix~\ref{app:ablations}).

\paragraph{Runtime checks and design intent.}
The model chooses the bindings between deliverables and workflow steps. The controller checks their execution, output forms, and quantities, but cannot determine whether a bound artifact is the one the brief requested or whether its content satisfies the constraints. Artifact evaluation remains necessary.

\paragraph{Task and interaction coverage.}
The benchmark uses English briefs and a scripted client for checkpoint questions. For a fair comparison, it includes only modalities that every compared system can generate: \system\ also generates 3D meshes and audio (grey in Figure~\ref{fig:architecture}), but not all compared systems do. We leave these modalities, other languages, and live client feedback to future evaluations.

\end{document}